%% file: template.tex
\documentclass{acmsiggraph}

\newcommand{\etal}{\mbox{\emph{et al.\ }}}

\newcommand{\eg}{\mbox{\emph{e.g.\ }}}
\newcommand{\etc}{\mbox{\emph{etc.\ }}}

\DeclareMathOperator*{\argmin}{arg\,min}

\usepackage{mathtools, cuted}

\title{Super-resolution Using Constrained Deep Texture Synthesis}

\author{Libin Sun\thanks{e-mail:lbsun@cs.brown.edu}\\Brown University
    \and
        James Hays\thanks{e-mail:hays@gatech.edu}\\Georgia Institute of Technology}
\pdfauthor{Libin Sun}


\setcopyright{none}
\begin{document}

%%% This is the ``teaser'' command, which puts an figure, centered, below 
%%% the title and author information, and above the body of the content.

% \teaser{
%   \includegraphics[height=1.5in]{figures/}
%   \caption{Spring Training 2009, Peoria, AZ.}
% }

\maketitle

\begin{abstract}
Hallucinating high frequency image details in single image super-resolution is a challenging task. Traditional super-resolution methods tend to produce oversmoothed output images due to the ambiguity in mapping between low and high resolution patches. We build on recent success in deep learning based texture synthesis and show that this rich feature space can facilitate successful transfer and synthesis of high frequency image details to improve the visual quality of super-resolution results on a wide variety of natural textures and images.
\end{abstract}

%\begin{CCSXML}
%<ccs2012>
%<concept>
%<concept_id>10010147.10010371.10010382</concept_id>
%<concept_desc>Computing methodologies~Image manipulation</concept_desc>
%<concept_significance>500</concept_significance>
%</concept>
%<concept>
%<concept_id>10010147.10010371.10010382.10010236</concept_id>
%<concept_desc>Computing methodologies~Computational photography</concept_desc>
%<concept_significance>300</concept_significance>
%</concept>
%</ccs2012>
%\end{CCSXML}

%\ccsdesc[500]{Computing methodologies~Image manipulation}
%\ccsdesc[300]{Computing methodologies~Computational photography}


% The next three commands are required, and insert the user-generated keywords, 
% The CCS concepts list, and the rights management text.
% Please make sure there is a blank line between each of these three commands.

\keywordlist

%\conceptlist

%\printcopyright

\input{introduction}
\input{related_work}

\input{method}

\input{results}
\input{discussion}

%\section*{Acknowledgements}

\bibliographystyle{acmsiggraph}
\bibliography{template}
\end{document}

%% file: introduction.tex
\section{Introduction}
\label{sec:intro}
Single image super-resolution (SISR) is a challenging problem due to its ill-posed nature--there exist many high resolution images (output) that could downsample to the same low resolution input image. Given moderate scaling factors, high contrast edges might warrant some extent of certainty in the high resolution output image, but smooth regions are impossible to recover unambiguously. As a result, most methods aim to intelligently hallucinate image details and textures while being faithful to the low resolution image~\cite{freeman_cga,Sun_sr,hacohen_sr_2010,sun_hays}. While recent state-of-the-art methods~\cite{sr_simplefunctions,aplusplus_sr,DeepSR_eccv2014,wang2015deep} are capable of delivering impressive performance in term of PSNR/SSIM metrics, the improvement in visual quality compared to earlier successful methods such as~\cite{yang_sr_2008} are not as apparent. In particular, the amount of image textural details are still lacking in these leading methods. We build on traditional and recent deep learning based texture synthesis approaches to show that reliable texture transfer can be achieved in the context of single image super-resolution and hallucination. 

Being able to model and represent natural image content is often a required first step towards recovering and hallucinating image details. Natural image models and priors have come a long way, from simple edge representations to more complex patch based models. Image restoration applications such as image super-resolution, deblurring, and denoising, share a similar philosophy in their respective framework to address the ill-posed nature of these tasks. A common strategy is to introduce image priors as a constraint in conjunction with the image formation model. Natural image content spans a broad range of spatial frequencies, and it is typically easy to constrain the restoration process to reliable recover information in the low frequency bands. These typically include smoothly varying regions without large gradients (edges, sky). In fact, a Gaussian or Laplacian prior would suit well for most image restoration task. This family of image priors have been shown to work in a variety of settings, in \cite{fergus_2006,Levin_2007_reflect,levin_2009,cho_2009,xu_jia_2010}, to name a few. More advanced prior models have also been developed such as FRAME~\cite{frame_1998}, the Fields of Experts model~\cite{foe_2009}, and the GMM model~\cite{Zoran2011}. It is known that the filters learned in these higher order models are essentially tuned low high-pass filters~\cite{weiss_freeman_2007}. As a result, no matter how these priors are formulated, they work under the same principle by penalizing high frequency image content, imposing the constraint that ``images should be smooth'' unless required by the image reconstruction constraint. When these priors are universally applied to every pixel location in the image, it is bound to yield over-smoothed output. But smoothness is just another form of blur, which is exactly what we are trying to avoid in the solution space in super-resolution.\\

To achieve sharpness in the upsampled image, successful methods usually learn a statistical mapping between low resolution (LR) and high resolution (HR) image patches. The mapping itself can be non-parametric~\cite{freeman_cga,jiabin_sr}, sparse coding~\cite{yang_sr_2008}, regression functions~\cite{kim_sr,sr_simplefunctions}, random forest~\cite{random_forest_sr}, and convolutional neural networks~\cite{DeepSR_eccv2014,wang2015deep,johnson_2016eccv}. There are pros and cons of both parametric and non-parametric representations. Parametric methods typically offer much faster performance at test time and produce higher PSNR/SSIM scores. But no matter how careful one engineers the loss function during training, the learned mapping will suffer from the inherent ambiguity in low to high resolution patch mapping (many-to-one), and end up with a conservative mapping to minimize loss (typically MMSE). This regression-towards-the-mean problem suppresses high frequency details in the HR output. Non-parametric methods are bound to the available example patch pairs in the training process, hence unable to synthesize new image content besides simple blending of patches. As a result, more artifacts can be found in the output image due to misalignment of image content in overlapping patches. However, non-parametric methods tend to be more aggressive in inserting image textures and details~\cite{hacohen_sr_2010,sun_hays}.

More recently, deep learning based approaches have been adopted with great success in many image restoration and synthesis tasks. The key is to use well-established deep networks as an extremely expressive feature space to achieve high quality results. In particular, a large body of work on image and texture synthesis have emerged and offer promising directions for single image super-resolution. By constraining the Gram matrix at different layers in a large pre-trained network, Gatys \etal showed that it is possible to synthesize a wide variety of natural image textures with almost photo-realistic quality~\cite{gatys_nips2015}. Augmenting the same constraint with another image similarity term, they showed that artistic styles can be transfered~\cite{gatys_styletransfer,gatys_transfer_control} from paintings to photos in the same efficient framework. Recent work~\cite{enhancenet,johnson_2016eccv} show that by training to minimize perceptual loss in the feature space, superior visual quality can be achieved for SISR. However, their success at synthesizing natural textures is still limited as shown in their examples.

In this work, we build on the same approach from~\cite{gatys_styletransfer} and adapt it handle SISR. We focus on synthesis and transfer aspect of natural image textures, and show that high frequency details can be reliably transfered and hallucinated from example images to render convincing HR output.

%% file: related_work.tex
\section{Related Work}
\label{sec:relatedwork}

\subsection{Single Image Super-resolution (SISR)}
Single image super-resolution is a long standing challenge in computer vision and image processing due to its extremely ill-posed nature. However, it has attracted much attention in recent research due to new possibilities introduced by big data and deep learning. Unlike traditional multi-frame SR, it is impossible to unambiguously restore high frequencies in a SISR framework. As a result, existing methods \emph{hallucinate} plausible image content by relying on carefully engineered constraints and optimization procedures.

Over the past decade, SISR methods have evolved from interpolation based and edge oriented methods to learning based approaches. Such methods learn a statistical model that maps low resolution (LR) patches to high resolution (HR) patches~\cite{yang_sr_2008,kim_sr,sr_simplefunctions,ann_sr,aplusplus_sr,random_forest_sr}, with deep-learning frameworks being the state-of-the-art~\cite{DeepSR_eccv2014,wang2015deep}. While these methods perform well in terms of PSNR/SSIM, high frequency details such as textures are still challenging to hallucinate because of the ambiguous mapping between LR and HR image patches. In this respect, non-parameteric patch-based methods have shown promising results~\cite{freeman_cga,sun_tappen,hacohen_sr_2010,sun_hays,jiabin_sr}. These methods introduce explicit spatial~\cite{freeman_cga} and contextual~\cite{sun_tappen,hacohen_sr_2010,sun_hays} constraints to insert appropriate image details using external example images. On the other hand, internal image statistics based methods have also shown great success~\cite{freedman_sr,glasner_2009,inplace_sr,michaeli_sr_2013,jiabin_sr}. These methods directly exploit self-similarity within and across spatial scales to achieve high quality results. 

More recently, new SISR approaches have emerged with an emphasis on synthesizing image details via deep networks to achieve better visual quality. Johnson \etal~\cite{johnson_2016eccv} show that the style transfer framework of~\cite{gatys_styletransfer} can be made real-time, and show that networks trained based on perceptual loss in the feature space can produce superior super-resolution results. Sajjadi \etal~\cite{enhancenet} consider the combination of several loss functions for training deep networks and compare their visual quality for SISR.

\subsection{Texture and Image synthesis}
In texture synthesis, the goal is to create an output image that matches the textural appearance of an input texture to minimize perceptual differences. Early attempts took a parametric approach~\cite{heeger_bergen,portilla} by matching statistical characteristics in a steerable pyramid. Non-parametric methods~\cite{debonet_1997,efros_leung_99,Efros01,graphcuttextures,wei_2000,kwatra_2005} completely sidestep statistical representation for textures, and synthesize textures by sampling pixels or patches in a nearest neighbor fashion. More recently, Gatys \etal~\cite{gatys_nips2015} propose Gram matrix based constraints in the rich and complex feature space of the well-known VGG network~\cite{vgg_2014}, and show impressive synthesized results on a diverse set of textures and images. This deep learning based approach shares many connections with earlier parametric models such as~\cite{heeger_bergen,portilla}, but relies on orders of magnitudes more parameters, hence is capable of more expressive representation of textures. 

Synthesizing an entire natural image from scratch is an extremely difficult task. Yet, recent advances in deep learning have shown promising success. Goodfellow \etal~\cite{gan} introduced the Generative Adversarial Network (GAN) to pair a discriminative and generative network together to train deep generative models capable of synthesizing realistic images. Follow-up works~\cite{lapgan,dcgan,ppgan} extended the GAN framework to improve the quality and resolution of generated images. However, the focus of this line of work has been to generate realistic images consistent with semantic labels such as object and image classes, in which low and mid level image features typically play a more crucial role, whereas the emphasis on high resolution image details and textures is not the primary goal.

\subsection{Image Style and Detail Transfer}
Many works exist in the domain of style and detail transfer between images. \cite{cg2real} enhance the realism of computer generated scenes by transfering color and texture details from real photographs. \cite{shih_2013} consider the problem of hallucinating time of day for a single photo by learning local affine transforms in a database of time-lapse videos. \cite{laffont_2014} utilize crowdsourcing to establish an annotated webcam database to facilitate transfering high level transient attributes among different scenes. Style transfer for specific image types such as portraits is also explored by~\cite{portrait_transfer}, in which multi-scale local transforms in a Laplacian pyramid are used to transfer contrast and color styling from exemplar professional portraits. 

More recently, \cite{gatys_styletransfer} propose a style transfer system using the 19-layer VGG network~\cite{vgg_2014}. The key constraint is to match the Gram matrix of numerous feature layers between the output image and a style image, while high level features of the output is matched that of a content image. In this way, textures of the style image is transfered to the output image as if painted over the content image, similar to Image Quilting~\cite{Efros01}. Drawing inspirations from texture synthesis methods, \cite{cnnmrf} propose to combine a MRF with CNN for image synthesis. This CNNMRF model adds additional layers in the network to enable resampling `neural patches', namely, each local window of the output image should be similar to some patch in the style image \emph{in feature space} in a nearest neighbor sense. This has the benefit of more coherent details should the style image be sufficiently representative of the content image. However, this copy-paste resampling mechanism is unable to synthesize new content. In addition, this method is prone to produce `washed out' artifacts due the blending/averaging of neural patches. This is a common problem to patch-based synthesis methods~\cite{Efros01,freeman_cga,kwatra_2005}. Other interesting deep learning based applications such as view synthesis~\cite{zhou2016viewsynthesis} and generative visual manipulation~\cite{zhu2016generative} have also been proposed. These methods allow us to better understand how to manipulate and transfer image details without sacrificing visual quality.

%% file: method.tex
\section{Method}
\label{sec:method}
\begin{figure*}
\begin{center}
\def\svgwidth{\textwidth}
   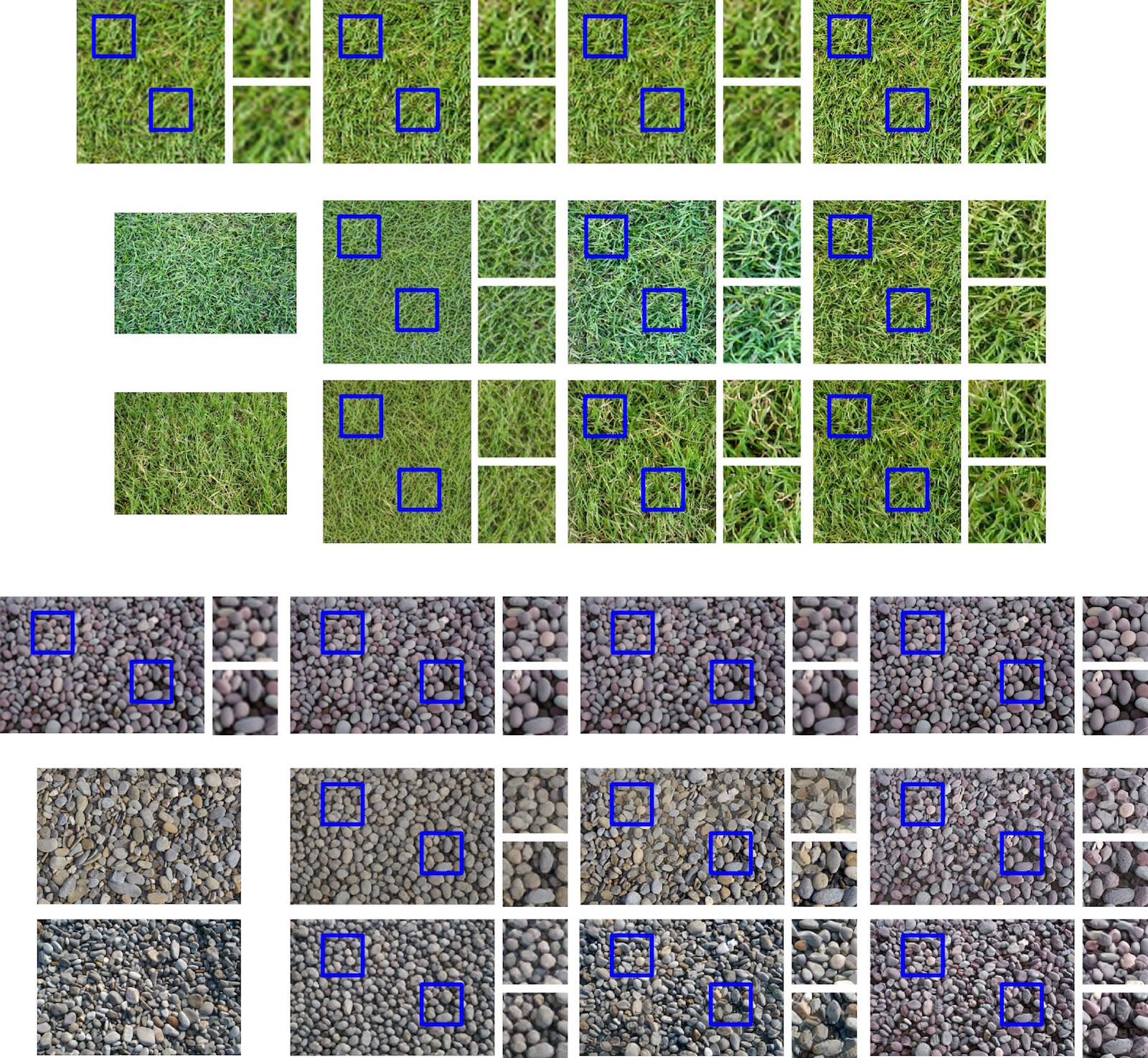
\end{center}
%\vspace{-15pt}
   \caption{A sample comparison of various algorithms applied to upsampling texture images for a factor of $\times 3$. Two example images are provided in both (a) and (b) for example-based approaches. It can be seen that the example image has significant impact on the appearance of the hallucinated details in the output images, indicating effectiveness of the texture transfer process.}
\label{fig:2styles}
\end{figure*}
Our method is based on \cite{gatys_styletransfer,gatys_nips2015}, which encodes feature correlations of an image in the VGG network via the Gram matrix. The VGG-Network is a 19-layer CNN that rivals human performance for the task of object recognition. This network consists of 16 convolutional layers, 5 pooling layers, and a series of fully connected layers for softmax classification.

A latent image $x$ is to be estimated given constraints such as content similarity and style similarity. We assume a style or example image $s$ is available for the transfer of appropriate textures from $s$ to $x$, and that $x$ should stay similar to a content image $c$ in terms of mid to high level image content. The feature space representations with the network are $X$, $S$ and $C$ respectively. At each layer $l$, a non-linear filter bank of $N_l$ filters is convolved with the previous layer's feature map to produce an encoding in the current layer, which can be stored in a feature matrix $X^l\in \mathcal{R}^{N_l\times M_l}$, where $M_l$ is the number of elements in the feature map (height times width). We use $X^l_{ij}$ to denote the activation of the $i^{th}$ filter at position $j$ in layer $l$ generated by image $x$. 

In \cite{gatys_styletransfer}, the goal is to solve for an image $x$ that is similar to a content image $c$ but takes on the style or textures of $s$. Specifically, the following objective function is minimized via gradient descent to solve for $x$:
\begin{equation}
\label{gatys_obj}
x = \argmin_x \left(\alpha E_{content}(c,x)+\beta E_{style}(s,x)\right)
\end{equation}
where $E_{content}$ is defined as:
\begin{equation}
E_{content}(c,x) = \frac{1}{2}\sum_l\sum_{ij}\left(C_{ij}^l - X_{ij}^l\right)^2
\end{equation}
The content similarity term is simply a $L_2$ loss given the difference between the feature map of the latent image in layer $l$ and the corresponding feature map from the content image. 

The definition of $E_{style}$ is based on the the $L_2$ loss between the Gram matrix of the latent image and the style image in a set of chosen layers. The Gram matrix encodes the correlations between the filter responses via the inner product of vectorized feature maps. Given a feature map $X^l$ for image $x$ in layer $l$, the Gram matrix $G(X^l)\in \mathcal{R}^{N_l\times M_l}$ has entries $G_{ij}^l=\sum_k X^l_{ik}X^l_{jk}$, where $i,j$ index through pairs of feature maps, and $k$ indexes through positions in each vectorized feature map. Then the style similarity component of the objective function is defined as:
\begin{equation}
E_{style}(s,x) = \sum_l\frac{w_l}{4N_l^2 M_l^2}\left(\sum_{i,j}\left(G(S^l)_{ij} - G(X^l)_{ij}\right)^2\right)
\end{equation}
where $w_l$ is a relative weight given to a particular layer $l$. The derivatives of the above energy terms can be found in~\cite{gatys_styletransfer}. To achieve best effect, the energy components are typically enforced over a set of layers in the network. For example, the content layer can be a single conv4\_2 layer, while the style layers can be over a larger set \{conv1\_1, conv2\_1, conv3\_1, conv4\_1, conv5\_1\} to allow consistent texture appearances across all spatial frequencies.

This feature space constraint has been shown to excel at representing natural image textures for texture synthesis, style transfer, and super-resolution. We introduce a few adaptations to the task of single image super-resolution and examine its effectiveness in terms of transfering and synthesizing natural textures.

\subsection{Basic Adaptation to SR}
The objective function in Equation~\ref{gatys_obj} consists of a content similarity term and a style term. The content term is analogous to the faithfulness term in SISR frameworks. The style term can be seen as a natural image prior derived from a single example image, which is assumed to represent the desired image statistics. A first step in our experiments is to replace the content similarity term $E_{content}$ with a faithfulness term $E_{faithfulness} = |G*x\downarrow_f - c|^2$, where $f$ is the downsampling factor, $G$ a Gaussian lowpass filter, and $c$ the low resolution input image that we would like to upsample. These variables associated with the downsampling process are assumed known a-priori (non-blind SR). In the subsequent discussion, we refer to this basic adaptation as \textbf{our global}, since the Gram matrix constraint is globally applied to the whole image. Formally, the \textbf{our global} method solves the following objective via gradient descent:
\begin{equation}
\label{global_obj}
x = \argmin_x \left(\alpha E_{faithfulness}(c,x)+\beta E_{style}(s,x)\right)
\end{equation}

We further make the following changes to the original setup:
\begin{itemize}
\item All processing is done in gray scale. The original work of \cite{gatys_styletransfer} computes the feature maps using RGB images. However, this requires strong similarity among color channel correlations between the example and input image, which is hard to achieve. For transfering artistic styles, this is not a problem. We drop the color information to allow better sharing of image statistics between the image pair.
\item We use the layers \{conv1\_1, pool1\_1, pool2\_1, pool3\_1, pool4\_1, pool5\_1\} to capture the statistics of the example image for better visual quality, as done in~\cite{gatys_nips2015}.
\end{itemize}

We show that the above setup, while simple and basic, is capable of transfering texture details reliably for a wide variety of textures (see Fig.\ref{fig:2styles} and Fig.\ref{fig:textures_various}), even if the textures are structured and regular (see Fig.\ref{fig:textures_regular}). However, for general natural scenes, this adaptation falls short and produces painterly artifacts or inappropriate image details for smooth image regions, because their global image statistics no longer matches each other.

\subsection{Local Texture Transfer via Masked Gram Matrices}
Natural images are complex in nature, usually consisting of a large number of segments and parts, some of which might contain homogeneous and stochastic textures. Clearly, globally matching image statistics for such complex scenes cannot be expected to yield good results. However, with carefully chosen local correspondences, we can selectively transfer image details by pairing image parts of the same or similar textures via two sets of binary masks $\{m_s^k\}_1^K$ and $\{m_x^k\}_1^K$. To achieve this, we introduce an outer summation to the $E_{style}$ term to loop over each corresponding pair of components in the masks (see Eq(\ref{eq:masked_eq})).
\begin{figure*}[!t]
\begin{align}
E_{stylelocal} & =  \sum_k E_{style}(s\otimes m^k_s,x\otimes m^k_x) =  \sum_k\sum_l\frac{w_l}{4N_l^2 |R^l_x(m_x^k)|^2}\left(\sum_{i,j}\left(G(S^l\otimes R_s^l(m_s^k))_{ij} - G(X^l\otimes R_x^l(m_x^k))_{ij}\right)^2\right) 
\label{eq:masked_eq}
\end{align}
\end{figure*}

In this setup, $R_x^l$ is an image resizing operator that resamples an image (a binary mask in this case) to the resolution of feature map $x^l$ using nearest neighbor interpolation. The normalization constant also reflects that we are aggregating image statistics over a subset of pixels in the images. The parameter $\beta$ from Eq.\ref{gatys_obj} is divided by the number of masks $K$ to ensure the same relative weight between $E_{faithfulness}$ and $E_{stylelocal}$. Note that these binary masks are not necessarily exclusive, namely, pixels can be explained by multiple masks if need be.

The sparse correspondences are non-trivial to obtain. We examine two cases for the correspondence via masks: manual masks, and automatic masks via the PatchMatch~\cite{patchmatch} algorithm.

\begin{figure}
\begin{center}
   \includegraphics[width=1\linewidth]{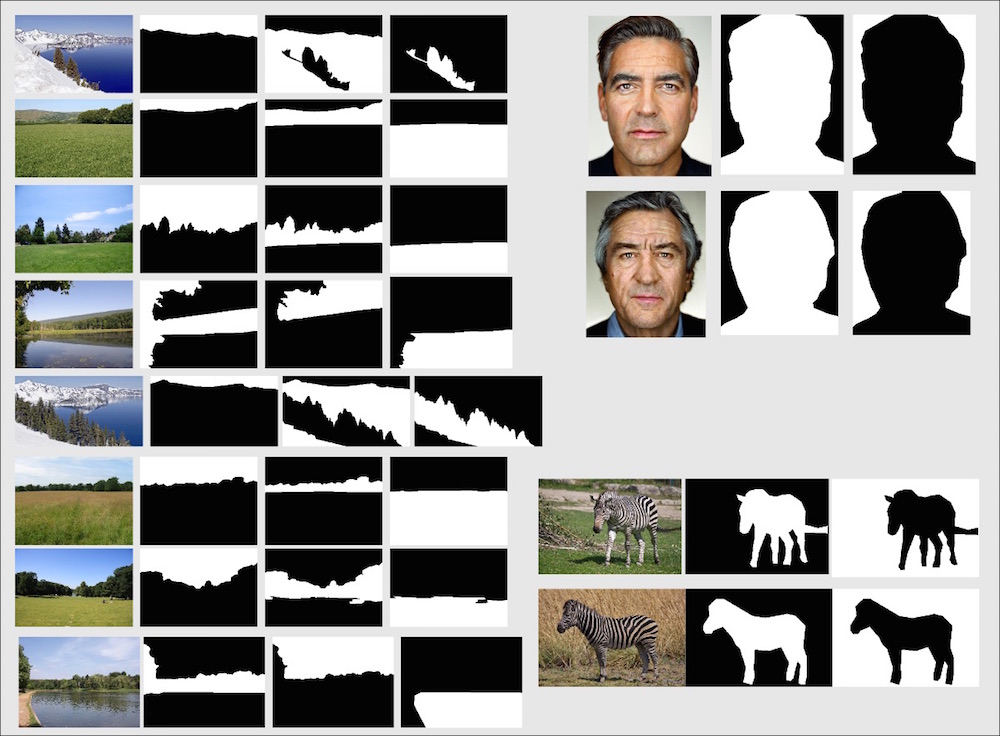}
\end{center}
%\vspace{-15pt}
   \caption{Sample images and their corresponding masks, each one is manually generated. }
   \label{fig:masks}
\end{figure}

\textbf{Manual Masks} For moderately simple scenes with large areas of homogeneous textures such as grass, trees, sky, \etc, we manually generate 2 to 3 masks per image at the full resolution to test out the local texture transfer. We refer to this setup as \textbf{our local manual}. A visualization of the images and masks can be found in Figure~\ref{fig:masks}.

\begin{figure*}
\begin{center}
\includegraphics[width=\textwidth]{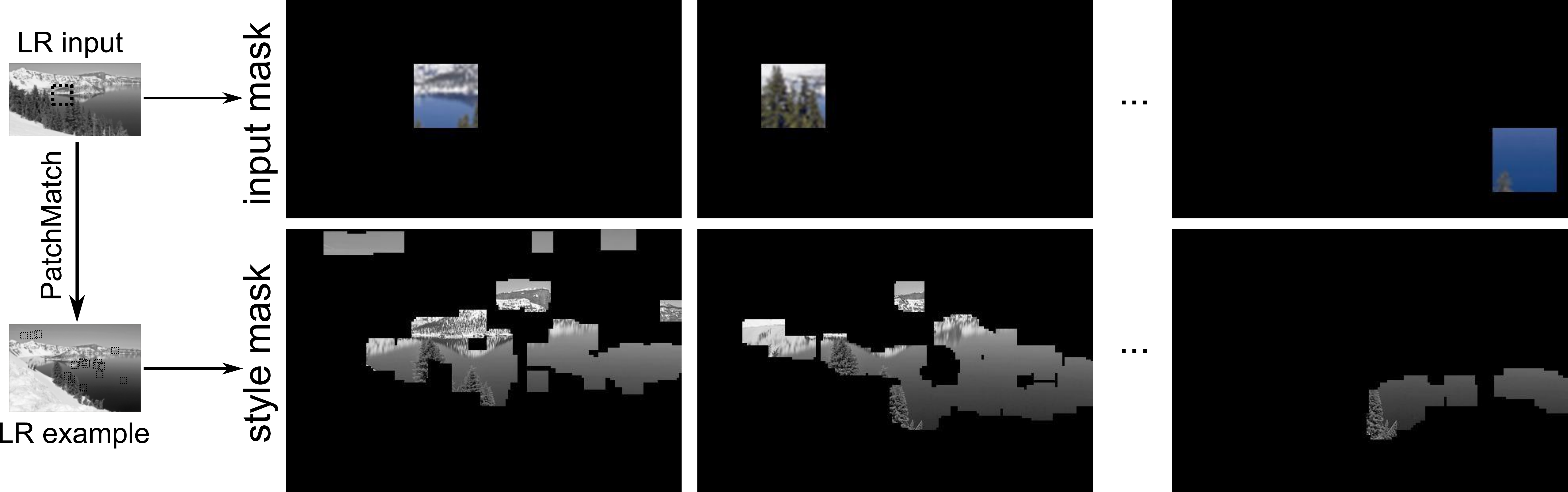}
\end{center}
%\vspace{-15pt}
   \caption{Visualization of the masks automatically generated using the PatchMatch algorithm. PatchMatch is applied to the low resolution grayscale input and example images to compute a dense correspondence. The HR output image is divided into cells, and all correspondences contained in the input cell are aggregated to form the example image mask. }
   \label{fig:patchmatch_vis}
\end{figure*}

\textbf{PatchMatch Masks} To automatically generate the masks, we apply the PatchMatch algorithm to the LR input image $c$ and a LR version of the style image $s$ after applying the same downsampling process used to generate $c$. Both images are grayscale. Once the nearest-neighbor field (NNF) is computed at the lower resolution, we divide the output image into cells and pool and dilate the interpolated offsets at the full resolution to form the mask pairs. Each $m_x^k$ contains a square cell of 1's, and its corresponding mask $m_s^k$ will be the union of numerous of binary patches. We refer to this variation as \textbf{our local}. A sample visualization is given in Figure~\ref{fig:patchmatch_vis}.

%% file: figures/2-style-new-small.pdf_tex
%% Creator: Inkscape 0.48.5, www.inkscape.org
%% Accompanies image file '2-style-new-small.pdf' (pdf, eps, ps)
\begingroup%
  \makeatletter%
  \providecommand\color[2][]{%
    \errmessage{(Inkscape) Color is used for the text in Inkscape, but the package 'color.sty' is not loaded}%
    \renewcommand\color[2][]{}%
  }%
  \providecommand\transparent[1]{%
    \errmessage{(Inkscape) Transparency is used (non-zero) for the text in Inkscape, but the package 'transparent.sty' is not loaded}%
    \renewcommand\transparent[1]{}%
  }%
  \providecommand\rotatebox[2]{#2}%
  \ifx\svgwidth\undefined%
    \setlength{\unitlength}{1600bp}%
    \ifx\svgscale\undefined%
      \relax%
    \else%
      \setlength{\unitlength}{\unitlength * \real{\svgscale}}%
    \fi%
  \else%
    \setlength{\unitlength}{\svgwidth}%
  \fi%
  \global\let\svgwidth\undefined%
  \global\let\svgscale\undefined%
  \makeatother%
  \begin{picture}(1,0.96230809)%
    \put(0,0.02){\includegraphics[width=\unitlength]{figures/2-style-new-small.jpg}}%
    \put(0.14158093,0.94987642){\color[rgb]{0,0,0}\makebox(0,0)[lb]{\smash{bicubic x3}}}%
    \put(0.54357342,0.77556115){\color[rgb]{0,0,0}\makebox(0,0)[lb]{\smash{Gatys transfer}}}%
    \put(0.74279082,0.77556115){\color[rgb]{0,0,0}\makebox(0,0)[lb]{\smash{our global transfer}}}%
    \put(0.56491802,0.94987645){\color[rgb]{0,0,0}\makebox(0,0)[lb]{\smash{SRCNN}}}%
    \put(0.35147053,0.77556115){\color[rgb]{0,0,0}\makebox(0,0)[lb]{\smash{CNNMRF}}}%
    \put(0.13090857,0.76488879){\color[rgb]{0,0,0}\makebox(0,0)[lb]{\smash{example image }}}%
    \put(0.76413555,0.94987645){\color[rgb]{0,0,0}\makebox(0,0)[lb]{\smash{groundtruth}}}%
    \put(0.36570051,0.94987645){\color[rgb]{0,0,0}\makebox(0,0)[lb]{\smash{ScSR}}}%
    \put(0.09533399,0.43048809){\color[rgb]{0,0,0}\makebox(0,0)[lb]{\smash{bicubic x3}}}%
    \put(0.59693511,0.43048809){\color[rgb]{0,0,0}\makebox(0,0)[lb]{\smash{SRCNN}}}%
    \put(0.83172723,0.43048809){\color[rgb]{0,0,0}\makebox(0,0)[lb]{\smash{groundtruth}}}%
    \put(0.35502814,0.43048809){\color[rgb]{0,0,0}\makebox(0,0)[lb]{\smash{ScSR}}}%
    \put(0.57203301,0.28107509){\color[rgb]{0,0,0}\makebox(0,0)[lb]{\smash{Gatys transfer}}}%
    \put(0.8103825,0.28107509){\color[rgb]{0,0,0}\makebox(0,0)[lb]{\smash{our global transfer}}}%
    \put(0.34435566,0.28107509){\color[rgb]{0,0,0}\makebox(0,0)[lb]{\smash{CNNMRF}}}%
    \put(0.07398932,0.28107509){\color[rgb]{0,0,0}\makebox(0,0)[lb]{\smash{example image}}}%
    \put(0.50444122,0.45183284){\color[rgb]{0,0,0}\makebox(0,0)[lb]{\smash{(a)}}}%
    \put(0.49732623,0.00359358){\color[rgb]{0,0,0}\makebox(0,0)[lb]{\smash{(b)}}}%
    \put(0.08821912,0.677){\color[rgb]{0,0,0}\rotatebox{90}{\makebox(0,0)[lb]{\smash{example 1}}}}%
    \put(0.08821912,0.51653939){\color[rgb]{0,0,0}\rotatebox{90}{\makebox(0,0)[lb]{\smash{example 2}}}}%
    \put(0.0206275,0.18427303){\color[rgb]{0,0,0}\rotatebox{90}{\makebox(0,0)[lb]{\smash{example 1}}}}%
    \put(0.0206275,0.05407027){\color[rgb]{0,0,0}\rotatebox{90}{\makebox(0,0)[lb]{\smash{example 2}}}}%
  \end{picture}%
\endgroup%

%% file: results.tex
\section{Experimental Results}
\label{sec:results}

\subsection{Baseline Methods}
For comparison, we first describe several baseline methods from recent literature on super-resolution and texture transfer, and compare to our methods. These baseline methods are representative of state-of-the-art performance in their respective tasks, and form the basis of comparison for Section~\ref{sec:finalresults}.

\textbf{ScSR}~\cite{yang_sr_2008,yang_sr_tip2010} is one of the most widely used methods for comparison in recent SISR literature. It is a sparse coding based approach, using a dictionary of 1024 atoms learned over a training set of 91 natural images. Sparse coding is a well studied framework for image reconstruction and restoration, in which the output signal is assumed to be a sparse linear activation of atoms from a learned dictionary. We use the Matlab implementation provided by the authors \footnote{We use the \texttt{Matlab} ScSR code package from \url{http://www.ifp.illinois.edu/~jyang29/codes/ScSR.rar}} as a baseline method for comparison.

\textbf{SRCNN}~\cite{DeepSR_eccv2014} is a CNN based SISR method that produces state-of-the-art performance for PSNR/SSIM measures among recent methods. It combines insights from sparse coding approaches and findings in deep learning. A 3-layer CNN architecture is proposed as an end-to-end system. We can view this representation as a giant non-linear regression system in neural space, mapping LR to HR image patches. For subsequent comparisons, we use the version of SRCNN learned from 5 million of $33\times 33$ subimages randomly sampled from ImageNet. The \texttt{Matlab} code package can be found on the author's website\footnote{We use the SRCNN code package from \url{http://mmlab.ie.cuhk.edu.hk/projects/SRCNN.html}}.

\textbf{Gatys}~\cite{gatys_styletransfer,gatys_nips2015} first consider reformulating the texture synthesis problem within a CNN framework. In both work, the VGG network is used for feature representation and modeling image space, and the correlation of feature maps at each layer is the key component in encoding textures and structures across spatial frequencies. The Gram matrix representation is compact and extremely effective at synthesizing a wide variety of textures~\cite{gatys_nips2015}. We use a Lasagne and Theano based implementation of~\cite{gatys_styletransfer} as a baseline method for comparison\footnote{Our implementation is adapted from the art style transfer recipe from Lasagne: \url{https://github.com/Lasagne/Recipes/tree/master/examples/styletransfer}}. 

\textbf{CNNMRF}~\cite{cnnmrf} address the loss of spatial information due to the Gram matrix representation by introducing an MRF style layer on top of the VGG hidden layers to constrain local similarity of \emph{neural patches}, where each local window in the output image feature map is constrained to be similar to the nearest neighbor in the corresponding layer of the style image feature maps. We use the \texttt{torch} based implementation from the authors\footnote{Chuan Li's CNNMRF implementation is available at: \url{https://github.com/chuanli11/CNNMRF}}.

To adapt the code from Gatys \etal and CNNMRF for our experiments, we upsample the LR input image bicubicly to serve as the content image. All other processing remain identical to their respective implementation.

We show a sample comparison of these methods in Figure~\ref{fig:2styles}, where a low resolution texture image is upsampled by a factor of 3. For the example based methods~\cite{gatys_styletransfer,cnnmrf} and ours, we provide two example images to test the algorithm's ability in transferring textures. Some initial observations can be made:
\begin{itemize}
\item ScSR~\cite{yang_sr_2008} and SRCNN~\cite{DeepSR_eccv2014} produce nearly identical results qualitatively, even though their model complexity is orders of magnitude apart. This represents half a decade of progress in the SISR literature.
\item CNNMRF~\cite{cnnmrf} produces painterly artifacts due to averaging in neural space. The highest frequencies among different color channels can be misaligned and appear as colored halos when zoomed in.
\item Our method produces convincing high frequency details while being faithful to the LR input. The effect of the example image can be clearly seen in the output image.
\end{itemize}

\input{results_show}

%% file: results_show.tex
\subsection{Comparison of Results}
\label{sec:finalresults}
In this section we showcase the performance of the algorithm variants \textbf{our global}, \textbf{our local} (PatchMatch based) and \textbf{our local manual} on a variety of textures and natural images. We also compare against leading methods in single-image super-resolution such as ScSR~\cite{yang_sr_2008} and SRCNN~\cite{DeepSR_eccv2014}, as well as deep learning based style transfer methods including \cite{gatys_styletransfer} and CNNMRF~\cite{cnnmrf}

\subsubsection{Test Data}
We collect a variety of images from the Internet including natural and man-made textures, regular textures, black and white patterns, text images, simple natural scenes consisting of 2 or 3 clearly distinguishable segments, and face images. These test images are collected specifically to test the texture transfer aspect of the algorithms. As a result, we do not evaluate performance of single image super-resolution in its traditional sense, namely, measuring PSNR and SSIM.

\subsubsection{Black and White Patterns}
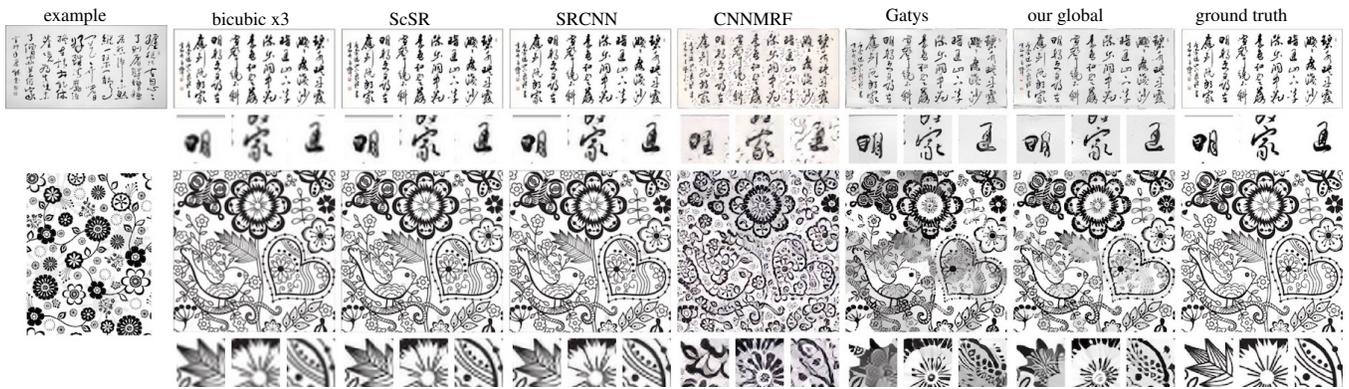
\begin{figure*}
\begin{center}
\def\svgwidth{\textwidth}
   \input{figures/textures_BWsketch.pdf_tex}
\end{center}
%\vspace{-15pt}
   \caption{Example comparisons on a Chinese text image (top) and black and white pattern image (bottom). Example based methods can hallucinate edges in interesting ways, but also produce biases in background intensity, copied from the example image. Other artifacts are also present. Best viewed electronically and zoomed in.}
   \label{fig:textures_bw}
\end{figure*}

The simplest test images are texts and black and white patterns. As shown in Figure~\ref{fig:textures_bw}, traditional SR algorithms do a decent job at sharpening strong edges, with SRCNN producing slightly less ringing artifacts than ScSR. As expected, the example based methods produce interesting hallucinated patterns based on the example image. CNNMRF yields considerable amount of artifacts due to averaging patches in neural space. Gatys and our global introduce a bias in background intensity but are capable of keeping the edges crisp and sharp. Much fine details and patterns are hallucinated for the bottom example.

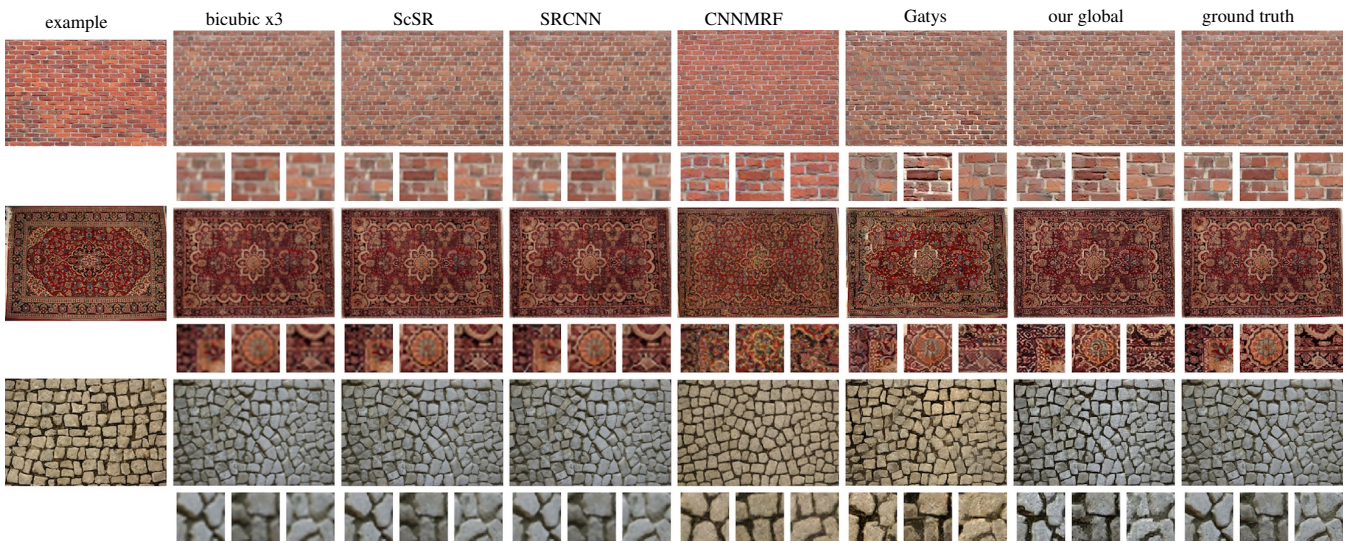
\begin{figure*}
\begin{center}
\def\svgwidth{\textwidth}
   \input{figures/textures_regular.pdf_tex}
\end{center}
%\vspace{-15pt}
   \caption{Example comparisons on regular textures. Best viewed electronically and zoomed in.}
\label{fig:textures_regular}
\end{figure*}

\subsubsection{Textures}
For homogeneous textures, most SISR methods simply cannot insert meaningful high frequency content besides edges. On the other hand, we see that the Gram matrix constraint from~\cite{gatys_styletransfer,gatys_nips2015} works extremely well because it is coercing image statistics across spatial frequencies in neural space, and ensuring that the output image match these statistics. However it is less effective when it comes to non-homogeneous image content such as edges and salient structures, or any type of image phenomena that is spatially unexchangeable. Finally, CNNMRF works reasonably well but still falls short in terms of realism. This is because linear blending of neural patches inevitably reduces high frequencies. Another artifact of this method is that this blending process can produce neural patches from the \emph{null space} of natural image patches, introducing colored halos and tiny rainbows when zoomed in.

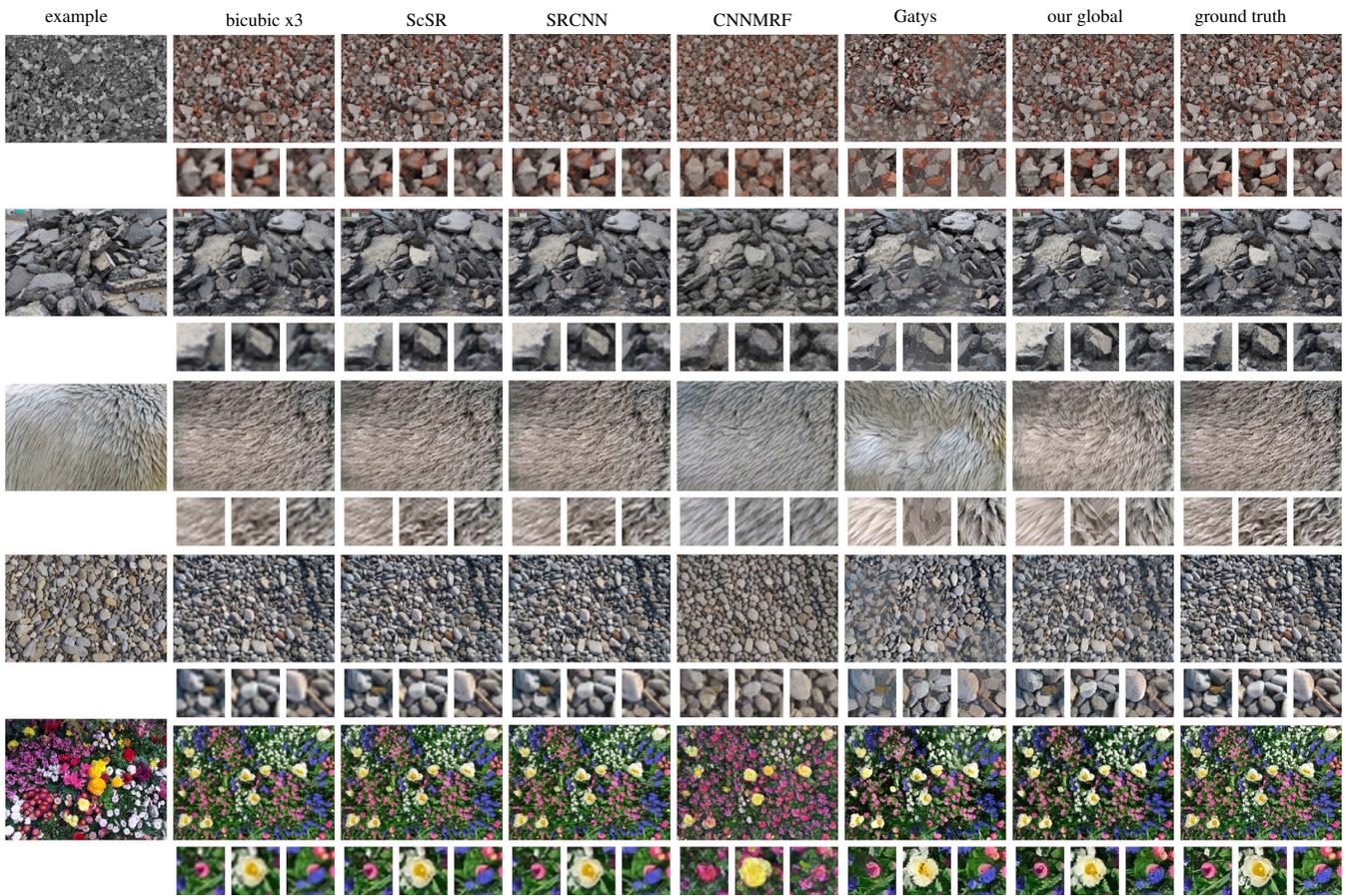
\begin{figure*}
\begin{center}
\def\svgwidth{\textwidth}
   \input{figures/textures_various.pdf_tex}
\end{center}
%\vspace{-15pt}
   \caption{Example comparisons on various types of textures. Best viewed electronically and zoomed in.}
\label{fig:textures_various}
\end{figure*}

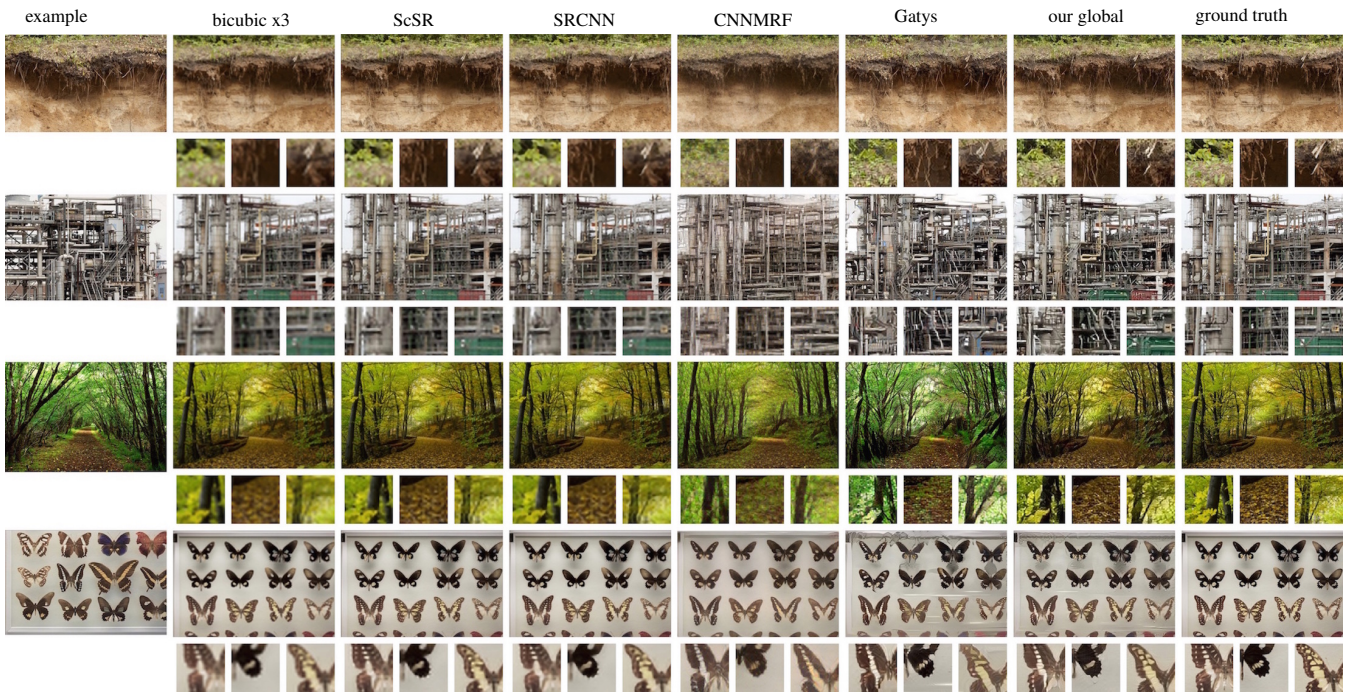
\begin{figure*}
\begin{center}
\def\svgwidth{\textwidth}
   \input{figures/textures_simpleimage.pdf_tex}
\end{center}
%\vspace{-15pt}
   \caption{Example comparisons on simple natural images. Best viewed electronically and zoomed in.}
\label{fig:textures_simpleimage}
\end{figure*}

The main benefits of the \textbf{our global} method are (1) better faithfulness to the input LR image, and (2) less color artifacts. The Gatys transfer baseline operates in RGB color space, hence any correlated color patterns from the style image will remain in the output image. However, the style image might might not represent the correct color correlation observed in the input image, \eg, blue vs yellow flowers against a background of green grass. Our global transfer method operates in gray scale, relaxing the correlation among color channels and allowing better sharing of image statistics. This relaxation helps bring out a more realistic output image, as shown in Figure~\ref{fig:textures_regular},~\ref{fig:textures_various},~\ref{fig:textures_simpleimage}.

\begin{figure*}
\begin{center}
\def\svgwidth{\textwidth}
   \input{figures/images_patchmatch.pdf_tex}
\end{center}
%\vspace{-15pt}
   \caption{Example comparisons on moderately complex natural images. CNNMRF, Gatys and `our local' consistently synthesize more high frequencies appropriate to the scene. CNNMRF and Gatys suffer from color artifacts due to mismatching colors between the example and the input image. CNNMRF also produces significant amount of color artifacts when viewed more closely, especially in smooth regions and near image borders. Gram matrix based methods such as Gatys and `our local' outperform other methods in terms of hallucinating image details, however also produce more artifacts in a few test cases. Best viewed electronically and zoomed in.}
   \label{fig:images_patchmatch}
\end{figure*}
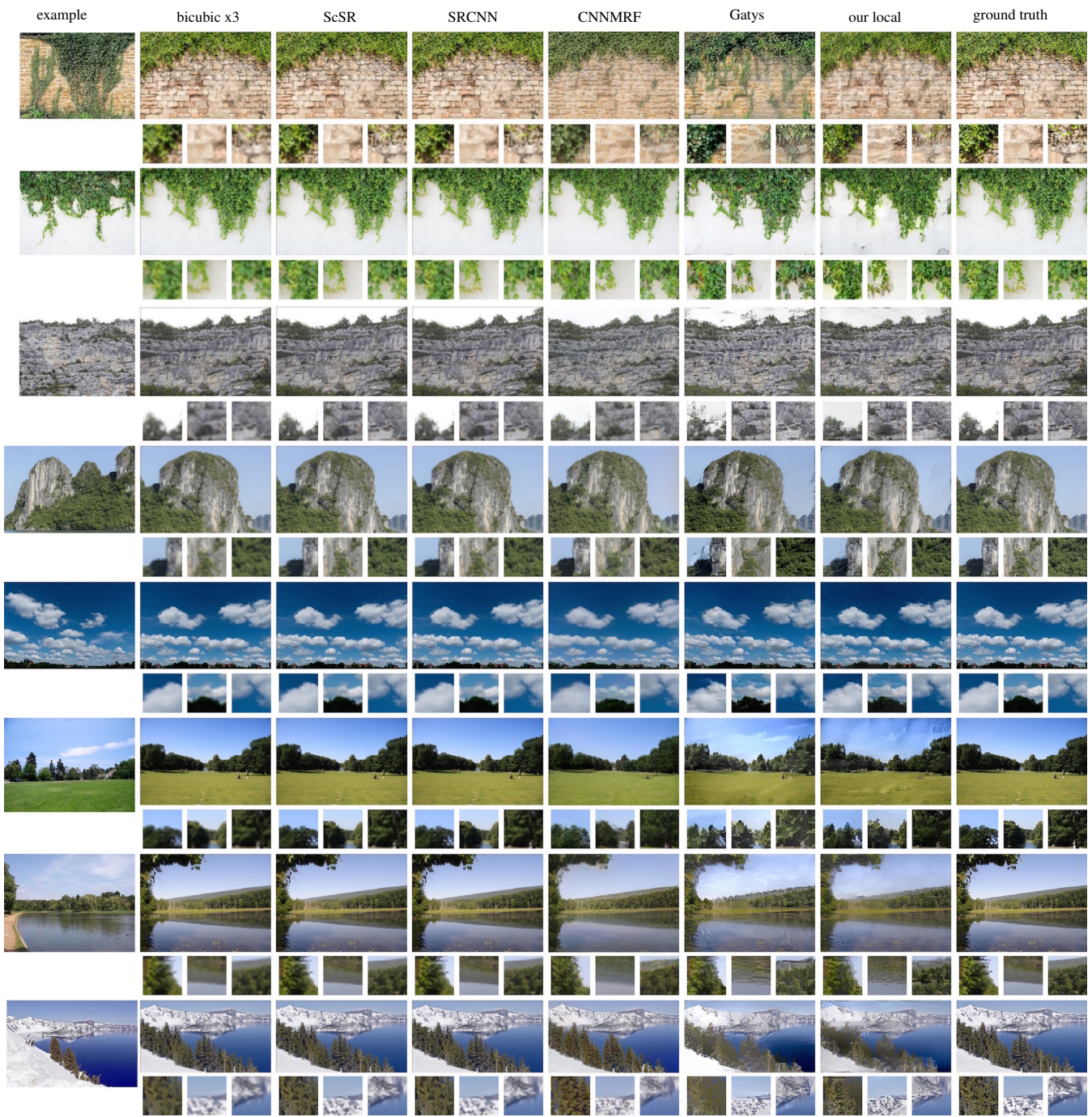

\begin{figure*}
\begin{center}
\def\svgwidth{\textwidth}
   \input{figures/images_manual.pdf_tex}
\end{center}
%\vspace{-15pt}
   \caption{Example comparisons on natural scenes with manually supplied masks. Best viewed electronically and zoomed in.}
   \label{fig:images_manual}
\end{figure*}
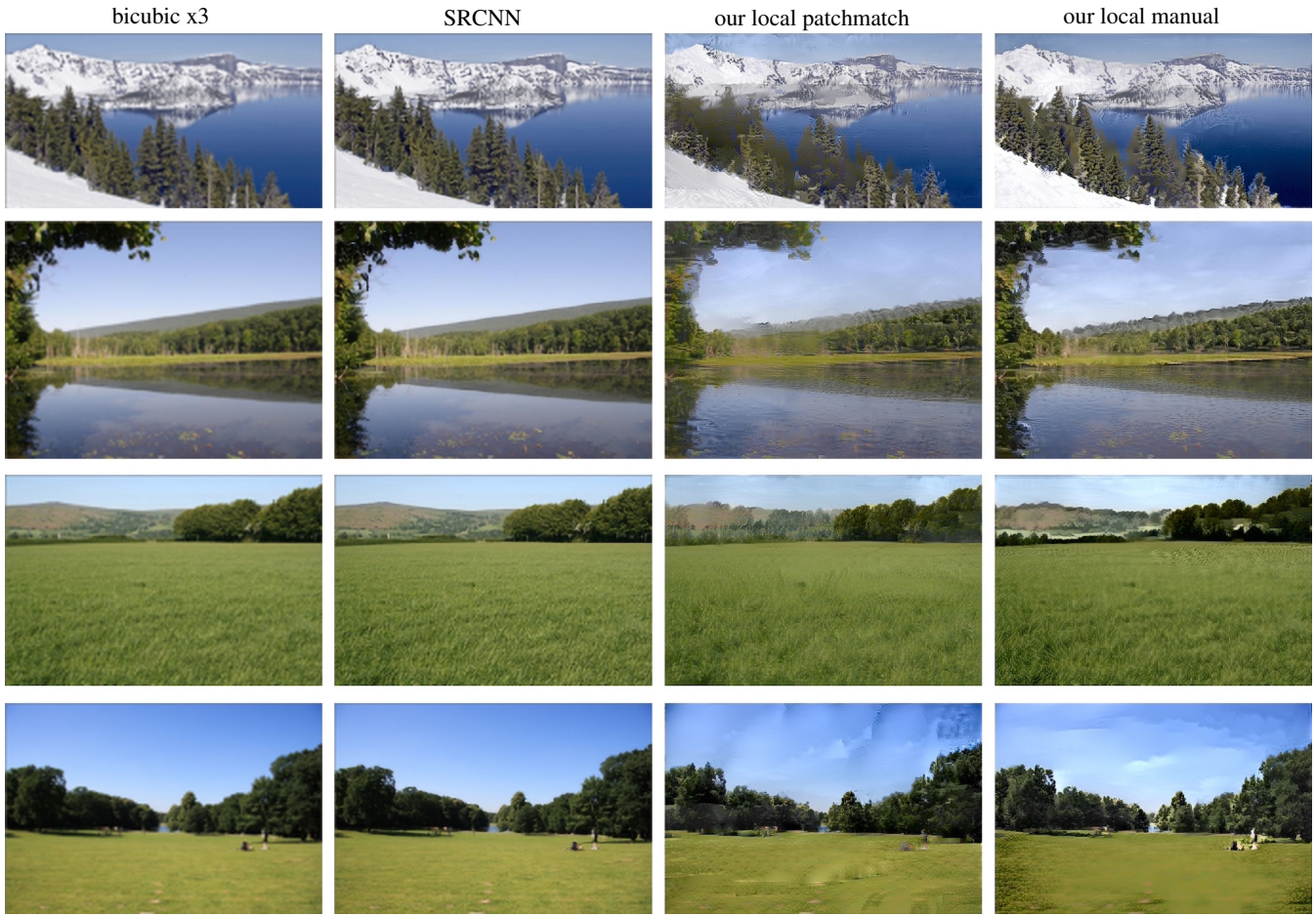

Comparisons on regular textures are shown in Figure~\ref{fig:textures_regular}. \textbf{our global} produces better details and color faithfulness, whereas traditional SISR methods do not appear too different from bicubic interpolation. Figure~\ref{fig:textures_various} shows results on numerous stochastic homogeneous textures. Example based methods exhibit strong influence from example images and can produce an output image visually different from the input, such as the fur image (third row). However, better details can be consistently observed throughout the examples. \textbf{Gatys} can be seen to produce a typical \emph{flat} appearance in color (\eg, rock, first row), this is because of the color processing constraint.

Going beyond homogeneous textures, we test these algorithms on simple natural images in Figure~\ref{fig:textures_simpleimage}. Realistic textures and details can be reasonably well hallucinated by \textbf{our global}, especially the roots in the roil (first row) and the patterns on the butterfly wings (bottom row). The pipes (second row) are synthesized well locally, however, the out output image becomes too `busy' when viewed globally. It is worth pointing out that CNNMRF essentially produces a painting for the forest image (third row), this is a clear example of the disadvantages of averaging/blending patches.

\subsubsection{Natural Scenes}
Natural images exhibit much more complexity than homogeneous textures, here we only consider scenarios where the image can be clearly divided into several types of textures, mostly homogeneous. In this way, we can better test the effectiveness of the algorithm's performance on synthesizing and hallucinating texture details. One complication that arises here is that texture transitions and borders represent extremely non-homogeneous statistics that is not easily handled by synthesis methods. Since the image now contains different types of statistics, we will apply our masked variants using PatchMatch masks and manual masks to these test images. To better deal with texture transitions, we dilate the manually generated masks slightly to include pixels near texture borders.

In Figure~\ref{fig:images_patchmatch}, all results under \textbf{our local} are generated using our PatchMatch based variant. These test images consist of moderately complex natural scenes. It can be seen that CNNMRF, Gatys and \textbf{our local} consistently synthesize more high frequencies appropriate to the scene, traditional SISR methods appear similar to bicubic interpolation. CNNMRF and Gatys suffer from color artifacts due to mismatching colors between the example and the input image. Again, CNNMRF produces significant amount of color artifacts when viewed more closely, especially in smooth regions and near image borders. Gram matrix based methods such as Gatys and \textbf{our local} outperform other methods in terms of hallucinating image details, however also produce more artifacts in a few test cases.

PatchMatch is far from perfect for generating the masks suitable for our application. This can be seen in many regions in the output images. For example, the trees in the pond image (second last row) is hallucinated by water textures towards the left, even the tree on the far left shows much water-like textures, clearly due to bad correspondences generated by PatchMatch. Similar artifacts can be seen in the crater lake image (last row). For natural scenes, our method is capable of opportunistically inserting appropriate textures, but cannot produce a perfect flaw-free output.

One would expect manually generated masks to be more suitable than PatchMatch masks. Although there are two drawbacks:
\begin{itemize}
\item The entire masked example region would participate in the Gram matrix computation, forcing the output image to take on the exemplar statistics, even though it might be undesirable. For example, when matching sky with slow intensity gradient with a flat sky region. PatchMatch offers more freedom in this regard, allowing certain regions to be completely discarded (in the example image).
\item Texture transitions are hard to account for. Even though we dilate the masks hoping to include the borders, the pyramid nature of the CNN architecture and pooling operations will eventually introduce boundary artifacts.
\end{itemize}

Figure~\ref{fig:images_manual} shows comparisons using our manually generally masks (c.r. Fig.~\ref{fig:masks}). Clearly, there is less low frequency artifacts in color biases. However, ringing artifacts become more prominent near texture transitions and image borders.

\subsection{Face Images}
\begin{figure}
\begin{center}
\def\svgwidth{\columnwidth}
   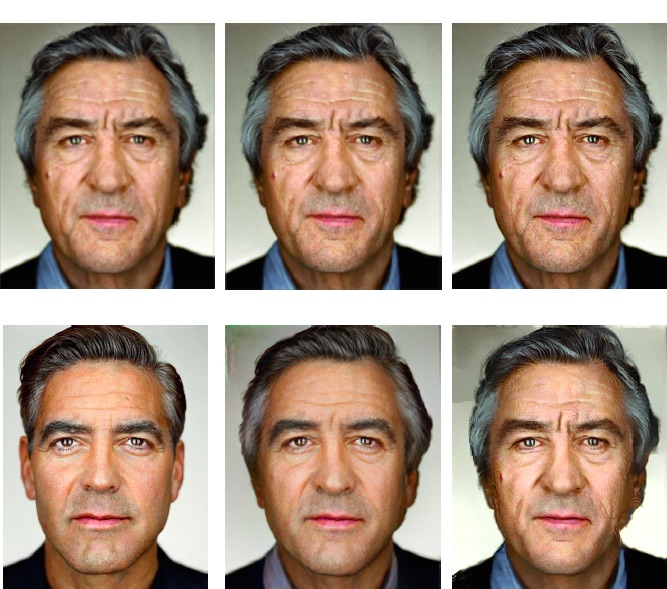
\end{center}
%\vspace{-15pt}
   \caption{Example comparisons on a portrait image. Our method is able to hallucinate appropriate details given the well-matched image statistics. Most noticeably, plausible details are successfully introduced to the eyebrows, hair, and eyes. CNNMRF produces decent amount of details as well, however, it makes the output image less recognizable as the person in the input image. Best viewed electronically.}
   \label{fig:images_deniro}
\end{figure}

Another interesting scenario is to test the algorithms on face images. When the example image is sufficiently close the input, such as in Figure~\ref{fig:images_deniro}, our method works well for hallucinating image details. In this particular example, the facial features in the output image remain similar to the input, and it is almost impossible to tell who the example image is given just the output. However, CNNMRF lacks the ability to synthesize new content (copy-paste in neural space) and its output is more of a blend between the input and example. The final output image somewhat falls into the `uncanny valley', and is almost unrecognizable as De Niro.

\begin{figure}
\begin{center}
\def\svgwidth{\columnwidth}
   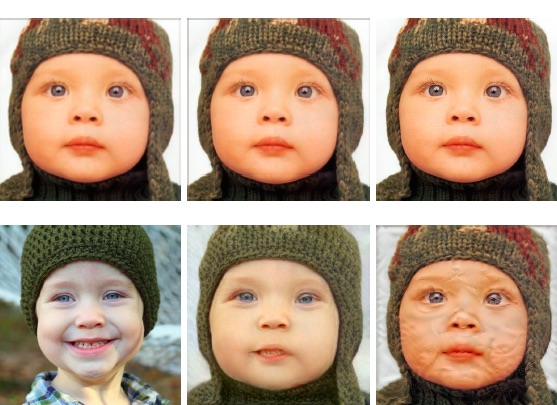
\end{center}
%\vspace{-15pt}
   \caption{Example comparisons on a face image. Our method fails due to mismatch in global image statistics. It is interesting to note that CNNMRF works extremely well for face images, however, it cannot insert image details not present in the example image. In this case, it cannot synthesize a closed mouth of the baby. Best viewed electronically.}
   \label{fig:images_baby}
\end{figure}

In Figure~\ref{fig:images_baby}, CNNMRF is able to produce a natural looking output with decent high frequency details except for the mouth region, since the example image does not contain the best source patches. On the other hand, our Gram matrix based method (our global setup) fails completely for the face region, only synthesizing details on parts of the hat, which happens to be homogeneous textures. This is because human faces are highly structured and far from textures.

%% file: figures/textures_BWsketch.pdf_tex
%% Creator: Inkscape 0.48.5, www.inkscape.org
%% Accompanies image file 'textures_BWsketch.pdf' (pdf, eps, ps)
\begingroup%
  \makeatletter%
  \providecommand\color[2][]{%
    \errmessage{(Inkscape) Color is used for the text in Inkscape, but the package 'color.sty' is not loaded}%
    \renewcommand\color[2][]{}%
  }%
  \providecommand\transparent[1]{%
    \errmessage{(Inkscape) Transparency is used (non-zero) for the text in Inkscape, but the package 'transparent.sty' is not loaded}%
    \renewcommand\transparent[1]{}%
  }%
  \providecommand\rotatebox[2]{#2}%
  \ifx\svgwidth\undefined%
    \setlength{\unitlength}{3312.00004281bp}%
    \ifx\svgscale\undefined%
      \relax%
    \else%
      \setlength{\unitlength}{\unitlength * \real{\svgscale}}%
    \fi%
  \else%
    \setlength{\unitlength}{\svgwidth}%
  \fi%
  \global\let\svgwidth\undefined%
  \global\let\svgscale\undefined%
  \makeatother%
  \begin{picture}(1,0.28953359)%
    \put(0,0){\includegraphics[width=\unitlength]{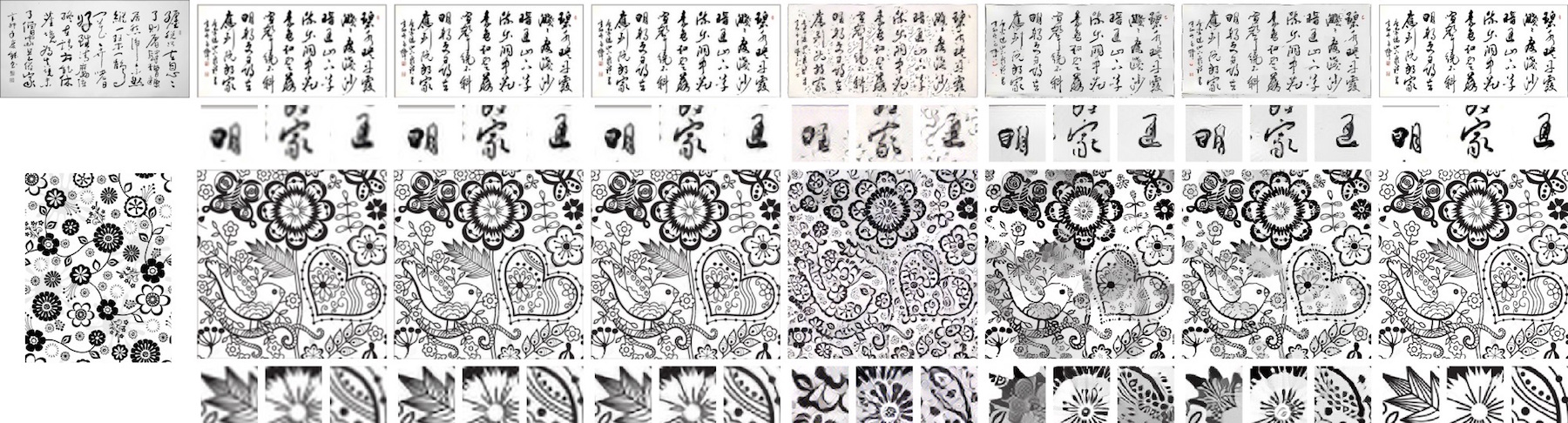}}%
    \put(0.15458937,0.27){\color[rgb]{0,0,0}\makebox(0,0)[lb]{\small{bicubic x3}}}%
    \put(0.65829322,0.27){\color[rgb]{0,0,0}\makebox(0,0)[lb]{\small{Gatys}}}%
    \put(0.76457342,0.27){\color[rgb]{0,0,0}\makebox(0,0)[lb]{\small{our global}}}%
    \put(0.41191642,0.27){\color[rgb]{0,0,0}\makebox(0,0)[lb]{\small{SRCNN}}}%
    \put(0.53,0.27){\color[rgb]{0,0,0}\makebox(0,0)[lb]{\small{CNNMRF}}}%
    %\put(0.01932367,0.17391304){\color[rgb]{0,0,0}\makebox(0,0)[lb]{\small{chinese2to1}}}%
    \put(0.02898551,0.27){\color[rgb]{0,0,0}\makebox(0,0)[lb]{\small{example}}}%
    %\put(0.01932367,0.01932367){\color[rgb]{0,0,0}\makebox(0,0)[lb]{\small{flowerBW14to8}}}%
    \put(0.28704333,0.27){\color[rgb]{0,0,0}\makebox(0,0)[lb]{\small{ScSR}}}%
    \put(0.89,0.27){\color[rgb]{0,0,0}\makebox(0,0)[lb]{\small{ground truth}}}%
  \end{picture}%
\endgroup%

%% file: figures/textures_regular.pdf_tex
%% Creator: Inkscape 0.48.5, www.inkscape.org
%% Accompanies image file 'textures_regular.pdf' (pdf, eps, ps)
\begingroup%
  \makeatletter%
  \providecommand\color[2][]{%
    \errmessage{(Inkscape) Color is used for the text in Inkscape, but the package 'color.sty' is not loaded}%
    \renewcommand\color[2][]{}%
  }%
  \providecommand\transparent[1]{%
    \errmessage{(Inkscape) Transparency is used (non-zero) for the text in Inkscape, but the package 'transparent.sty' is not loaded}%
    \renewcommand\transparent[1]{}%
  }%
  \providecommand\rotatebox[2]{#2}%
  \ifx\svgwidth\undefined%
    \setlength{\unitlength}{3311.99995117bp}%
    \ifx\svgscale\undefined%
      \relax%
    \else%
      \setlength{\unitlength}{\unitlength * \real{\svgscale}}%
    \fi%
  \else%
    \setlength{\unitlength}{\svgwidth}%
  \fi%
  \global\let\svgwidth\undefined%
  \global\let\svgscale\undefined%
  \makeatother%
  \begin{picture}(1,0.3940385)%
    \put(0,0){\includegraphics[width=\unitlength]{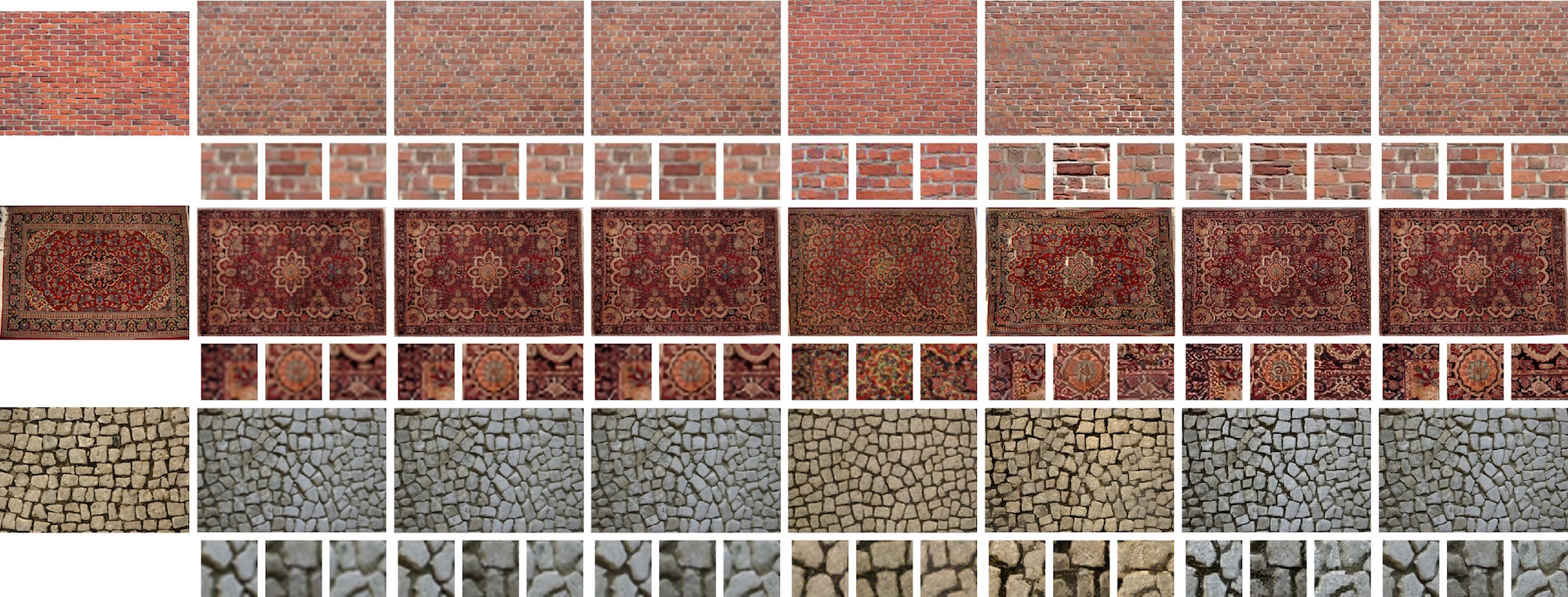}}%
    \put(0.15,0.385){\color[rgb]{0,0,0}\makebox(0,0)[lb]{\small{bicubic x3}}}%
    \put(0.67198828,0.385){\color[rgb]{0,0,0}\makebox(0,0)[lb]{\small{Gatys}}}%
    \put(0.78,0.385){\color[rgb]{0,0,0}\makebox(0,0)[lb]{\small{our global}}}%
    \put(0.4,0.385){\color[rgb]{0,0,0}\makebox(0,0)[lb]{\small{SRCNN}}}%
    \put(0.523,0.385){\color[rgb]{0,0,0}\makebox(0,0)[lb]{\small{CNNMRF}}}%
    \put(0.03,0.38){\color[rgb]{0,0,0}\makebox(0,0)[lb]{\small{example}}}
    \put(0.895,0.385){\color[rgb]{0,0,0}\makebox(0,0)[lb]{\small{ground truth}}}%
    %\put(0.01690821,0.27536232){\color[rgb]{0,0,0}\makebox(0,0)[lb]{\small{brickwall1to6}}}%
    %\put(0.03623188,0.14009662){\color[rgb]{0,0,0}\makebox(0,0)[lb]{\small{rug2to1}}}%
    %\put(0.03623188,0.02415459){\color[rgb]{0,0,0}\makebox(0,0)[lb]{\small{floors84to58}}}%
    \put(0.29,0.385){\color[rgb]{0,0,0}\makebox(0,0)[lb]{\small{ScSR}}}%
  \end{picture}%
\endgroup%

%% file: figures/textures_various.pdf_tex
%% Creator: Inkscape 0.48.5, www.inkscape.org
%% Accompanies image file 'textures_various.pdf' (pdf, eps, ps)
\begingroup%
  \makeatletter%
  \providecommand\color[2][]{%
    \errmessage{(Inkscape) Color is used for the text in Inkscape, but the package 'color.sty' is not loaded}%
    \renewcommand\color[2][]{}%
  }%
  \providecommand\transparent[1]{%
    \errmessage{(Inkscape) Transparency is used (non-zero) for the text in Inkscape, but the package 'transparent.sty' is not loaded}%
    \renewcommand\transparent[1]{}%
  }%
  \providecommand\rotatebox[2]{#2}%
  \ifx\svgwidth\undefined%
    \setlength{\unitlength}{3252.9875bp}%
    \ifx\svgscale\undefined%
      \relax%
    \else%
      \setlength{\unitlength}{\unitlength * \real{\svgscale}}%
    \fi%
  \else%
    \setlength{\unitlength}{\svgwidth}%
  \fi%
  \global\let\svgwidth\undefined%
  \global\let\svgscale\undefined%
  \makeatother%
  \begin{picture}(1,0.65016437)%
    \put(0,0){\includegraphics[width=\unitlength]{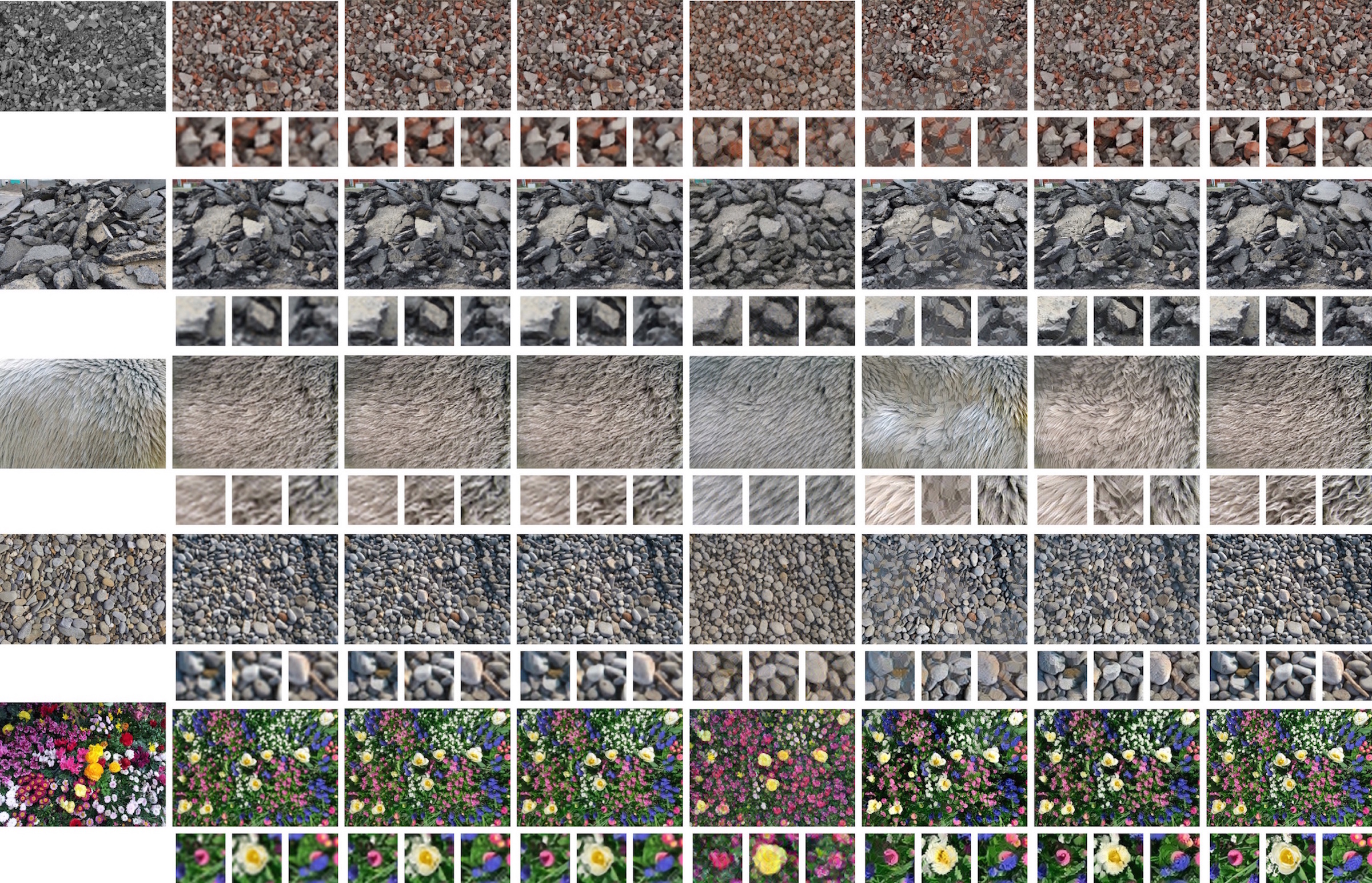}}%
    \put(0.16500846,0.65){\color[rgb]{0,0,0}\makebox(0,0)[lb]{\small{bicubic x3}}}%
    \put(0.665,0.65){\color[rgb]{0,0,0}\makebox(0,0)[lb]{\small{Gatys}}}%
    \put(0.78,0.65){\color[rgb]{0,0,0}\makebox(0,0)[lb]{\small{our global}}}%
    \put(0.405,0.65){\color[rgb]{0,0,0}\makebox(0,0)[lb]{\small{SRCNN}}}%
    \put(0.53,0.65){\color[rgb]{0,0,0}\makebox(0,0)[lb]{\small{CNNMRF}}}%
    %\put(0.04520079,0.53363014){\color[rgb]{0,0,0}\makebox(0,0)[lb]{\small{debris2to1}}}%
    \put(0.03,0.65){\color[rgb]{0,0,0}\makebox(0,0)[lb]{\small{example}}}%
%    \put(0.04520079,0.39591059){\color[rgb]{0,0,0}\makebox(0,0)[lb]{\small{debris4to3}}}%
%    \put(0.04520079,0.14014569){\color[rgb]{0,0,0}\makebox(0,0)[lb]{\small{gravel3to13}}}%
%    \put(0.04520079,0.00242614){\color[rgb]{0,0,0}\makebox(0,0)[lb]{\small{flower15to17}}}%
    \put(0.3,0.65){\color[rgb]{0,0,0}\makebox(0,0)[lb]{\small{ScSR}}}%
%    \put(0.0550379,0.26802814){\color[rgb]{0,0,0}\makebox(0,0)[lb]{\small{fur4to1}}}%
    \put(0.89,0.65){\color[rgb]{0,0,0}\makebox(0,0)[lb]{\small{ground truth}}}%
  \end{picture}%
\endgroup%

%% file: figures/textures_simpleimage.pdf_tex
%% Creator: Inkscape 0.48.5, www.inkscape.org
%% Accompanies image file 'textures_simpleimage.pdf' (pdf, eps, ps)
\begingroup%
  \makeatletter%
  \providecommand\color[2][]{%
    \errmessage{(Inkscape) Color is used for the text in Inkscape, but the package 'color.sty' is not loaded}%
    \renewcommand\color[2][]{}%
  }%
  \providecommand\transparent[1]{%
    \errmessage{(Inkscape) Transparency is used (non-zero) for the text in Inkscape, but the package 'transparent.sty' is not loaded}%
    \renewcommand\transparent[1]{}%
  }%
  \providecommand\rotatebox[2]{#2}%
  \ifx\svgwidth\undefined%
    \setlength{\unitlength}{3200bp}%
    \ifx\svgscale\undefined%
      \relax%
    \else%
      \setlength{\unitlength}{\unitlength * \real{\svgscale}}%
    \fi%
  \else%
    \setlength{\unitlength}{\svgwidth}%
  \fi%
  \global\let\svgwidth\undefined%
  \global\let\svgscale\undefined%
  \makeatother%
  \begin{picture}(1,0.50711526)%
    \put(0,0){\includegraphics[width=\unitlength]{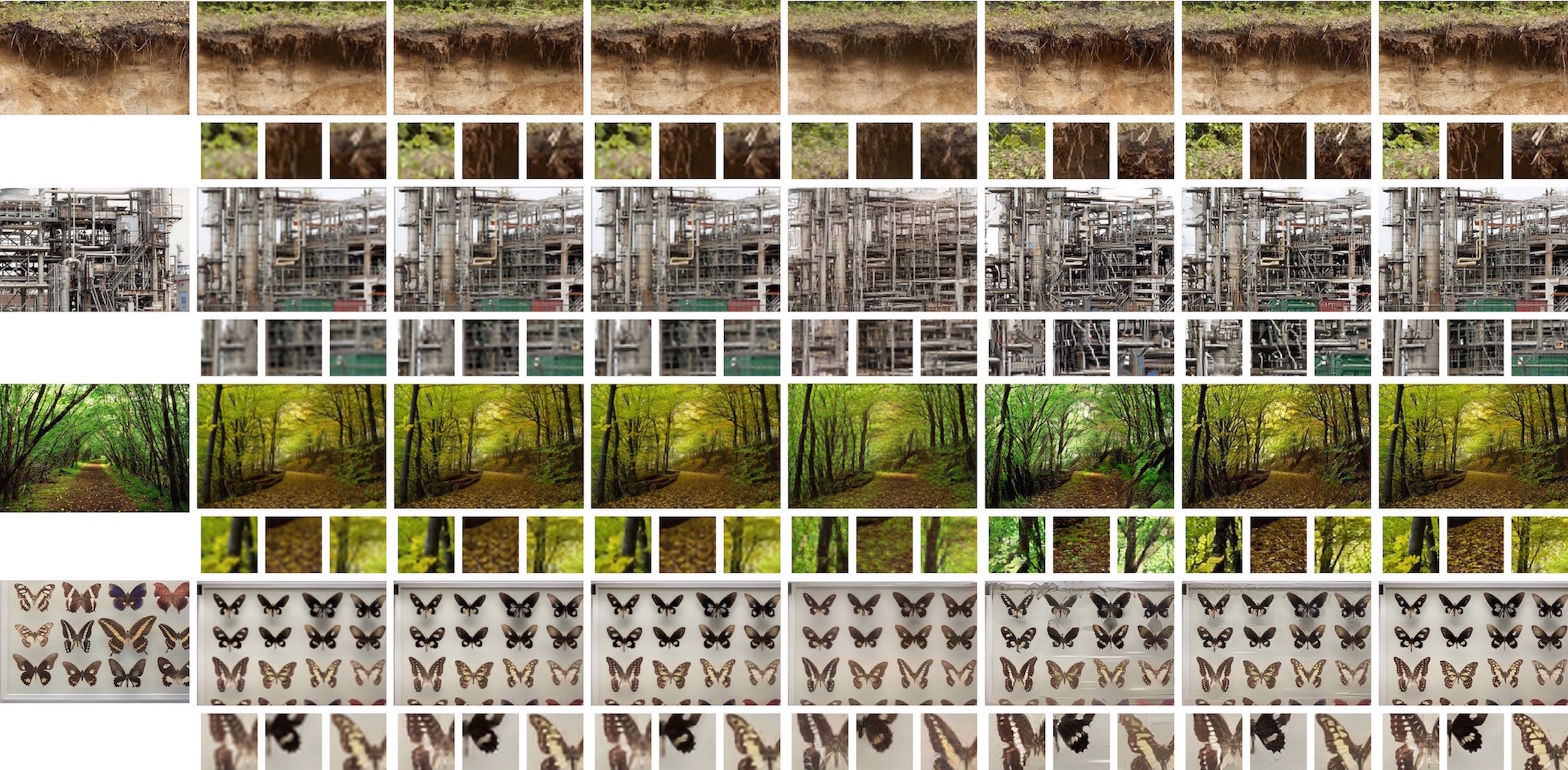}}%
    \put(0.15500003,0.49752542){\color[rgb]{0,0,0}\makebox(0,0)[lb]{\small{bicubic x3}}}%
    \put(0.66500003,0.49752542){\color[rgb]{0,0,0}\makebox(0,0)[lb]{\small{Gatys}}}%
    \put(0.78,0.49752542){\color[rgb]{0,0,0}\makebox(0,0)[lb]{\small{our global}}}%
    \put(0.41,0.49752542){\color[rgb]{0,0,0}\makebox(0,0)[lb]{\small{SRCNN}}}%
    \put(0.53,0.49752542){\color[rgb]{0,0,0}\makebox(0,0)[lb]{\small{CNNMRF}}}%
    \put(0.015,0.49752542){\color[rgb]{0,0,0}\makebox(0,0)[lb]{\small{example}}}%
    %\put(0.03,0.39252542){\color[rgb]{0,0,0}\makebox(0,0)[lb]{\small{soil1to2}}}%
    %\put(0.03,0.26752548){\color[rgb]{0,0,0}\makebox(0,0)[lb]{\small{plant3to4}}}%
    %\put(0.025,0.13252548){\color[rgb]{0,0,0}\makebox(0,0)[lb]{\small{forest3to1}}}%
    \put(0.29,0.49752542){\color[rgb]{0,0,0}\makebox(0,0)[lb]{\small{ScSR}}}%
    \put(0.89,0.49852542){\color[rgb]{0,0,0}\makebox(0,0)[lb]{\small{ground truth}}}%
    %\put(0.02,0.00252539){\color[rgb]{0,0,0}\makebox(0,0)[lb]{\small{butterfly2to1}}}%
  \end{picture}%
\endgroup%

%% file: figures/images_patchmatch.pdf_tex
%% Creator: Inkscape 0.48.5, www.inkscape.org
%% Accompanies image file 'images_patchmatch.pdf' (pdf, eps, ps)
\begingroup%
  \makeatletter%
  \providecommand\color[2][]{%
    \errmessage{(Inkscape) Color is used for the text in Inkscape, but the package 'color.sty' is not loaded}%
    \renewcommand\color[2][]{}%
  }%
  \providecommand\transparent[1]{%
    \errmessage{(Inkscape) Transparency is used (non-zero) for the text in Inkscape, but the package 'transparent.sty' is not loaded}%
    \renewcommand\transparent[1]{}%
  }%
  \providecommand\rotatebox[2]{#2}%
  \ifx\svgwidth\undefined%
    \setlength{\unitlength}{3312.00003119bp}%
    \ifx\svgscale\undefined%
      \relax%
    \else%
      \setlength{\unitlength}{\unitlength * \real{\svgscale}}%
    \fi%
  \else%
    \setlength{\unitlength}{\svgwidth}%
  \fi%
  \global\let\svgwidth\undefined%
  \global\let\svgscale\undefined%
  \makeatother%
  \begin{picture}(1,1.01085636)%
    \put(0,0){\includegraphics[width=\unitlength]{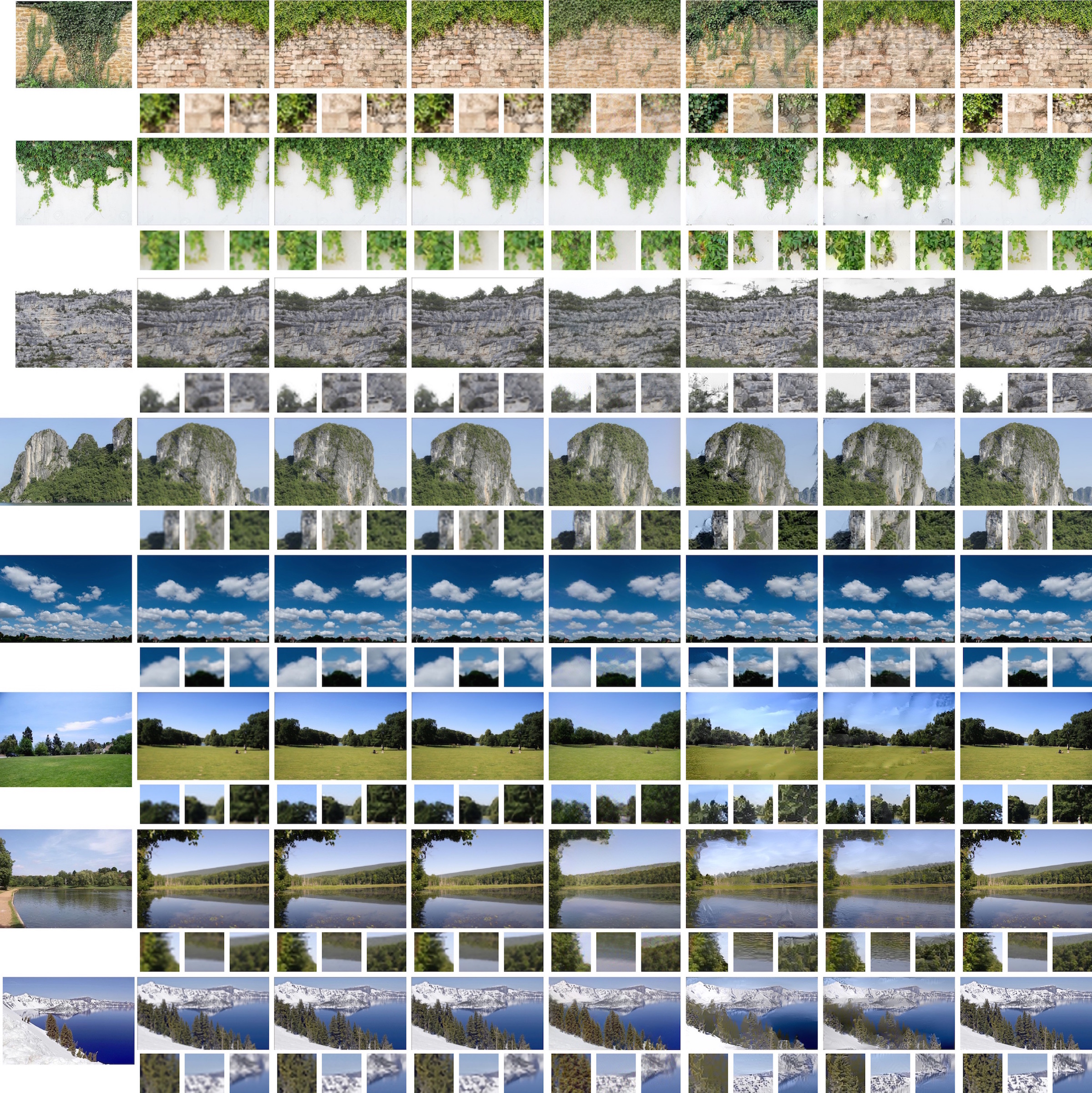}}%
    \put(0.15942028,1.01){\color[rgb]{0,0,0}\makebox(0,0)[lb]{\small{bicubic x3}}}%
    \put(0.67,1.01){\color[rgb]{0,0,0}\makebox(0,0)[lb]{\small{Gatys}}}%
    \put(0.78,1.01){\color[rgb]{0,0,0}\makebox(0,0)[lb]{\small{our local}}}%
    \put(0.41,1.01){\color[rgb]{0,0,0}\makebox(0,0)[lb]{\small{SRCNN}}}%
    \put(0.53,1.01){\color[rgb]{0,0,0}\makebox(0,0)[lb]{\small{CNNMRF}}}%
    \put(0.03,1.01){\color[rgb]{0,0,0}\makebox(0,0)[lb]{\small{example}}}%
     \put(0.29468598,1.01){\color[rgb]{0,0,0}\makebox(0,0)[lb]{\small{ScSR}}}%
    \put(0.895,1.01){\color[rgb]{0,0,0}\makebox(0,0)[lb]{\small{ground truth}}}%
    
    %\put(0.05314009,0.89613527){\color[rgb]{0,0,0}\makebox(0,0)[lb]{\small{ivy4to3}}}%
    %\put(0.05314009,0.77053141){\color[rgb]{0,0,0}\makebox(0,0)[lb]{\small{ivy1to2}}}%
    %\put(0.03381642,0.6352657){\color[rgb]{0,0,0}\makebox(0,0)[lb]{\small{mountain1to2}}}%
    %\put(0.03864734,0.50966184){\color[rgb]{0,0,0}\makebox(0,0)[lb]{\small{rock9to11}}}%
    %\put(0.04347826,0.384058){\color[rgb]{0,0,0}\makebox(0,0)[lb]{\small{sky2to1}}}%
    %\put(0.02173912,0.2584541){\color[rgb]{0,0,0}\makebox(0,0)[lb]{\small{grassfield9to10}}}%
    %\put(0.02173912,0.12801932){\color[rgb]{0,0,0}\makebox(0,0)[lb]{\small{pond2012_1to0}}}%
    %\put(0.01690821,0.01207729){\color[rgb]{0,0,0}\makebox(0,0)[lb]{\small{craterlake4to5}}}%
  \end{picture}%
\endgroup%

%% file: figures/images_manual.pdf_tex
%% Creator: Inkscape 0.48.5, www.inkscape.org
%% Accompanies image file 'images_manual.pdf' (pdf, eps, ps)
\begingroup%
  \makeatletter%
  \providecommand\color[2][]{%
    \errmessage{(Inkscape) Color is used for the text in Inkscape, but the package 'color.sty' is not loaded}%
    \renewcommand\color[2][]{}%
  }%
  \providecommand\transparent[1]{%
    \errmessage{(Inkscape) Transparency is used (non-zero) for the text in Inkscape, but the package 'transparent.sty' is not loaded}%
    \renewcommand\transparent[1]{}%
  }%
  \providecommand\rotatebox[2]{#2}%
  \ifx\svgwidth\undefined%
    \setlength{\unitlength}{1648bp}%
    \ifx\svgscale\undefined%
      \relax%
    \else%
      \setlength{\unitlength}{\unitlength * \real{\svgscale}}%
    \fi%
  \else%
    \setlength{\unitlength}{\svgwidth}%
  \fi%
  \global\let\svgwidth\undefined%
  \global\let\svgscale\undefined%
  \makeatother%
  \begin{picture}(1,0.69850901)%
    \put(0,0){\includegraphics[width=\unitlength]{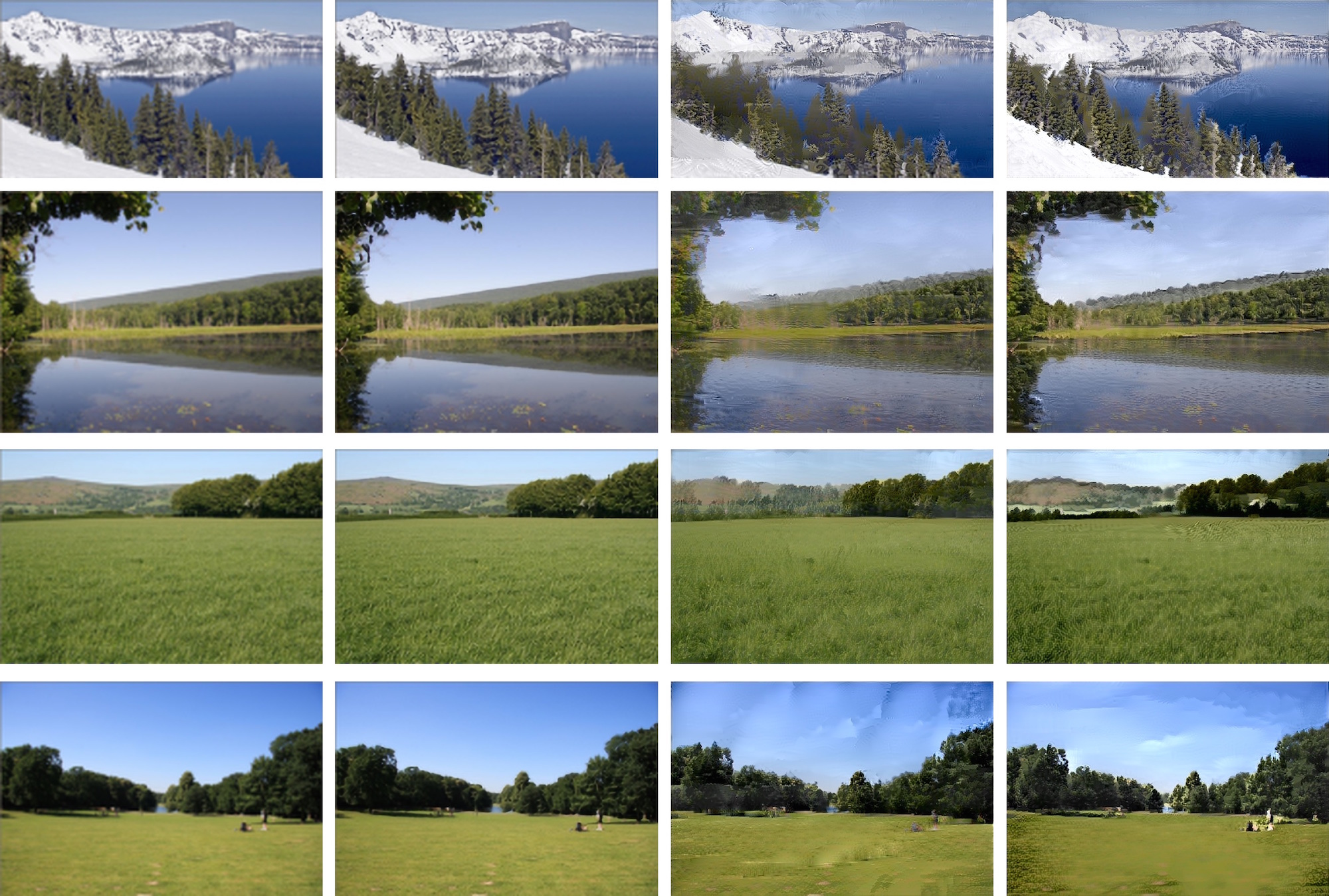}}%
    \put(0.08183079,0.68176144){\color[rgb]{0,0,0}\makebox(0,0)[lb]{\smash{bicubic x3}}}%
    \put(0.33564492,0.68037449){\color[rgb]{0,0,0}\makebox(0,0)[lb]{\smash{SRCNN}}}%
    \put(0.543,0.68037447){\color[rgb]{0,0,0}\makebox(0,0)[lb]{\smash{our local patchmatch}}}%
    \put(0.81,0.68176144){\color[rgb]{0,0,0}\makebox(0,0)[lb]{\smash{our local manual}}}%
  \end{picture}%
\endgroup%

%% file: figures/images_portrait_deniro.pdf_tex
%% Creator: Inkscape 0.48.5, www.inkscape.org
%% Accompanies image file 'images_portrait_deniro.pdf' (pdf, eps, ps)
\begingroup%
  \makeatletter%
  \providecommand\color[2][]{%
    \errmessage{(Inkscape) Color is used for the text in Inkscape, but the package 'color.sty' is not loaded}%
    \renewcommand\color[2][]{}%
  }%
  \providecommand\transparent[1]{%
    \errmessage{(Inkscape) Transparency is used (non-zero) for the text in Inkscape, but the package 'transparent.sty' is not loaded}%
    \renewcommand\transparent[1]{}%
  }%
  \providecommand\rotatebox[2]{#2}%
  \ifx\svgwidth\undefined%
    \setlength{\unitlength}{1229.91081543bp}%
    \ifx\svgscale\undefined%
      \relax%
    \else%
      \setlength{\unitlength}{\unitlength * \real{\svgscale}}%
    \fi%
  \else%
    \setlength{\unitlength}{\svgwidth}%
  \fi%
  \global\let\svgwidth\undefined%
  \global\let\svgscale\undefined%
  \makeatother%
  \begin{picture}(1,0.88373556)%
    \put(0,0){\includegraphics[width=\unitlength]{figures/images_portrait_deniro.jpg}}%
    \put(0.09561668,0.865){\color[rgb]{0,0,0}\makebox(0,0)[lb]{\smash{bicubic x3}}}%
    \put(0.44853425,0.865){\color[rgb]{0,0,0}\makebox(0,0)[lb]{\smash{SRCNN}}}%
    \put(0.75536253,0.865){\color[rgb]{0,0,0}\makebox(0,0)[lb]{\smash{ground truth}}}%
    \put(0.44165804,0.4049537){\color[rgb]{0,0,0}\makebox(0,0)[lb]{\smash{CNNMRF}}}%
    \put(0.10490888,0.4051395){\color[rgb]{0,0,0}\makebox(0,0)[lb]{\smash{example}}}%
    \put(0.7246983,0.4051395){\color[rgb]{0,0,0}\makebox(0,0)[lb]{\smash{our local manual}}}%
  \end{picture}%
\endgroup%

%% file: figures/images_portrait_baby.pdf_tex
%% Creator: Inkscape 0.48.5, www.inkscape.org
%% Accompanies image file 'images_portrait_baby.pdf' (pdf, eps, ps)
\begingroup%
  \makeatletter%
  \providecommand\color[2][]{%
    \errmessage{(Inkscape) Color is used for the text in Inkscape, but the package 'color.sty' is not loaded}%
    \renewcommand\color[2][]{}%
  }%
  \providecommand\transparent[1]{%
    \errmessage{(Inkscape) Transparency is used (non-zero) for the text in Inkscape, but the package 'transparent.sty' is not loaded}%
    \renewcommand\transparent[1]{}%
  }%
  \providecommand\rotatebox[2]{#2}%
  \ifx\svgwidth\undefined%
    \setlength{\unitlength}{1232bp}%
    \ifx\svgscale\undefined%
      \relax%
    \else%
      \setlength{\unitlength}{\unitlength * \real{\svgscale}}%
    \fi%
  \else%
    \setlength{\unitlength}{\svgwidth}%
  \fi%
  \global\let\svgwidth\undefined%
  \global\let\svgscale\undefined%
  \makeatother%
  \begin{picture}(1,0.73019481)%
    \put(0,0){\includegraphics[width=\unitlength]{figures/images_portrait_baby.jpg}}%
    \put(0.0974026,0.70779221){\color[rgb]{0,0,0}\makebox(0,0)[lb]{\smash{bicubic x3}}}%
    \put(0.44155844,0.7038961){\color[rgb]{0,0,0}\makebox(0,0)[lb]{\smash{SRCNN}}}%
    \put(0.42857143,0.33506494){\color[rgb]{0,0,0}\makebox(0,0)[lb]{\smash{CNNMRF}}}%
    \put(0.11038961,0.33766234){\color[rgb]{0,0,0}\makebox(0,0)[lb]{\smash{example}}}%
    \put(0.75324675,0.70779221){\color[rgb]{0,0,0}\makebox(0,0)[lb]{\smash{ground truth}}}%
    \put(0.77272727,0.33766234){\color[rgb]{0,0,0}\makebox(0,0)[lb]{\smash{our global}}}%
  \end{picture}%
\endgroup%

%% file: discussion.tex
\section{Discussion and Future Work}
\label{sec:discussion}
Recent works on the texture synthesis aspect of single image super-resolution provide a promising direction that complements existing methods which perform well in traditional image quality metrics. We have shown that deep architectures can provide the appropriate constraints in its rich feature space to model natural image content, especially textures. We have shown that the Gram matrix constraint from~\cite{gatys_nips2015} can be easily adapted to achieve realistic transfer of high frequency details for wide variety of natural textures and images. With sparse spatial correspondences, more localized transfer of textures can be achieved to handle moderately complex natural scenes. However, it is non-trivial to handle texture transitions by matching statistics in neural space. Non-homogeneous textures, edges, and objects in natural images are also challenging to handle by this framework. Future work may focus on combining texture and object synthesis with traditional SISR approach for edge handling in a more unified framework.